\documentclass[twoside,11pt]{article}

\usepackage{jmlr2e}
\usepackage{bm}
\usepackage{dsfont}

\usepackage[english]{babel}

\usepackage[noframe]{showframe}
\usepackage{framed}
\usepackage{lipsum}
\definecolor{color0}  {RGB}{174,225,254} 
\definecolor{color1}  {RGB}{220,227,248} 
\definecolor{color2}  {RGB}{28,130,185} 
\definecolor{color3}  {RGB}{255,253,250} 
\newenvironment{svgraybox}{%
	\MakeFramed{\advance\hsize-\width \FrameRestore\FrameRestore}}%
{\endMakeFramed}
\definecolor{shadecolor}{gray}{0.75}

\usepackage{tcolorbox}
\usepackage{extarrows}

\usepackage{graphicx}  
\usepackage{float}  
\usepackage{subfigure}  

\usepackage{algorithm}
\usepackage{algpseudocode}
\usepackage{amsmath}
\usepackage{graphics}
\usepackage{epsfig}
\usepackage[colorlinks]{hyperref}

\usepackage{blkarray}

\usepackage[framemethod=tikz]{mdframed}

\newcommand{\mdframecolorNote}{gray!10}
\newcommand{\mdframehidelineNote}{true}

\newcommand{\mdframecolorSkip}{yellow!10}

\usepackage{hyperref}
\hypersetup{
    colorlinks,
    citecolor=black,
    filecolor=black,
    linkcolor=black,
    urlcolor=black
}

\newcommand{\real}{\mathbb{R}}

\newcommand{\leadto}{\qquad\underrightarrow{ \text{leads to} }\qquad}
\newcommand{\leadtosmall}{\,\,\,\,\underrightarrow{ \text{leads to} }\,\,\,\,}

\mathchardef\mhyphen="2D

\newcommand{\diag}{\mathrm{diag}}

\newcommand{\mathcalV}{\mathcal{V}}

\newcommand{\trans}[1]{\ensuremath{#1^{ \top}}}

\newcommand{\cspace}{\mathcal{C}}
\newcommand{\nspace}{\mathcal{N}}

\newcommand{\bzero}{\boldsymbol{0}}

\newcommand{\balpha}{\boldsymbol\alpha}
\newcommand{\bbeta}{\boldsymbol\beta}

\newcommand{\boldeta}{\boldsymbol\eta}

\newcommand{\bmu}{\boldsymbol\mu}

\newcommand{\bSigma}{\boldsymbol\Sigma}

\newcommand{\bLambda}{\boldsymbol\Lambda}

\newcommand{\ba}{\bm{a}}
\newcommand{\bA}{\bm{A}}

\newcommand{\bB}{\bm{B}}
\newcommand{\bc}{\bm{c}}
\newcommand{\bC}{\bm{C}}
\newcommand{\bd}{\bm{d}}
\newcommand{\bD}{\bm{D}}
\newcommand{\be}{\bm{e}}
\newcommand{\bE}{\bm{E}}

\newcommand{\bG}{\bm{G}}

\newcommand{\bH}{\bm{H}}

\newcommand{\bI}{\bm{I}}

\newcommand{\bL}{\bm{L}}

\newcommand{\bM}{\bm{M}}
\newcommand{\bn}{\bm{n}}

\newcommand{\bp}{\bm{p}}
\newcommand{\bP}{\bm{P}}
\newcommand{\bq}{\bm{q}}
\newcommand{\bQ}{\bm{Q}}
\newcommand{\br}{\bm{r}}
\newcommand{\bR}{\bm{R}}

\newcommand{\bu}{\bm{u}}
\newcommand{\bU}{\bm{U}}
\newcommand{\bv}{\bm{v}}
\newcommand{\bV}{\bm{V}}

\newcommand{\bx}{\bm{x}}
\newcommand{\bX}{\bm{X}}
\newcommand{\by}{\bm{y}}
\newcommand{\bY}{\bm{Y}}
\newcommand{\bz}{\bm{z}}
\newcommand{\bZ}{\bm{Z}}

\jmlrheading{1}{2021}{1-48}{4/00}{10/00}{Jun Lu}

\ShortHeadings{Revisit the Fundamental Theorem of Linear Algebra}{Jun Lu}
\firstpageno{1}

\begin{document}

\title{Revisit the Fundamental Theorem of Linear Algebra}

\author{
\begin{center}
	\name Jun Lu \\ 
	\email jun.lu.locky@gmail.com
\end{center}
   }


\maketitle

\begin{abstract}
This survey is meant to provide an introduction to the fundamental theorem of linear algebra and the theories behind them. Our goal is to give a rigorous introduction to the readers with prior exposure to linear algebra. Specifically, we provide some details and proofs of some results from \citep{strang1993fundamental}.
We then describe the fundamental theorem of linear algebra from different views and find the properties and relationships behind the views. The fundamental theorem of linear algebra is essential in many fields, such as electrical engineering, computer science, machine learning and, deep learning.
This survey is primarily a summary of purpose, significance of important theories behind it. 

The sole aim of this survey is to give a self-contained introduction to concepts and mathematical tools in theory behind the fundamental theorem of linear algebra and rigorous analysis in order to seamlessly introduce its properties in four subspaces in subsequent sections. However, we clearly realize our inability to cover all the useful and interesting results and given the paucity of scope to present this discussion, e.g., the separated analysis of the (orthogonal) projection matrices. We refer the reader to literature in the field of linear algebra for a more detailed introduction to the related fields. Some excellent examples include \citep{rose1982linear, strang1993introduction, trefethen1997numerical, strang2019linear, strang2021every}.



\end{abstract}


\tableofcontents
\newpage

\section{Introduction and Background}
The readers with enough background of matrix analysis can skip this section.
In all cases, scalars will be denoted in a non-bold font possibly with subscripts (e.g., $a$, $\alpha$, $\alpha_i$). We will use bold face lower case letters possibly with subscripts to denote vectors (e.g., $\bmu$, $\bx$, $\bx_n$, $\bz$) and
bold face upper case letters possibly with subscripts to denote matrices (e.g., $\bA$, $\bL_j$). The $i$-th element of a vector $\bz$ will be denoted by $\bz_i$ in bold font. The $i$-th row and $j$-th column element of matrix $\bA$ will be denoted by $\bA_{ij}$. Furthermore, it will be helpful to utilize the \textbf{Matlab-style notation}, the $i$-th row to $j$-th row and $k$-th column to $m$-th column submatrix of matrix $\bA$ will be denoted by $\bA_{i:j,k:m}$. And in all cases, vectors are formulated in a column rather than in a row. A row vector will be denoted by a transpose of a column vector such as $\ba^\top$. A specific column vector with values are split by symbol $";"$, e.g., $\bx=[1;2;3]$ is a column vector in $\real^3$. Similarly, a specific row vector with values are split by symbol $","$, e.g., $\by=[1,2,3]$ is a row vector with 3 values. Further, a column vector can be denoted by the transpose of a row vector e.g., $\by=[1,2,3]^\top$ is a column vector.

The transpose of a matrix $\bA$ will be denoted by $\bA^\top$ and its inverse will be denoted by $\bA^{-1}$ . We will denote the $p \times p$ identity matrix by $\bI_p$.  A vector or matrix of all zeros will be denoted by a bold face zero $\bzero$ whose size should be clear from context, or we denote $\bzero_p$ to be the vector of all zeros with $p$ entries.

\begin{definition}[Eigenvalue]
	Given any vector space $E$ and any linear map $A: E \rightarrow E$, a scalar $\lambda \in K$ is called an eigenvalue, or proper value, or characteristic value of $\bA$ if there is some nonzero vector $\bu \in E$ such that
	\begin{equation*}
		\bA \bu = \lambda \bu.
	\end{equation*}
\end{definition}

\begin{definition}[Eigenvector]
	A vector $\bu \in E$ is called an eigenvector, or proper vector, or characteristic vector of $\bA$ if $\bu \neq 0$ and if there is some $\lambda \in K$ such that
	\begin{equation*}
		\bA \bu  = \lambda \bu,
	\end{equation*}
	where the scalar $\lambda$ is then an eigenvalue. And we say that $\bu$ is an eigenvector associated with $\lambda$.
\end{definition}



\begin{definition}[Linearly Independent]
	A set of vectors $\{\ba_1, \ba_2, \cdots, \ba_n\}$ is called linearly independent if there is no combination can get $x_1\ba_1+x_2\ba_2+\cdots+x_n\ba_n=0$ except all $x_i$'s are zero. An equivalent definition is that $\ba_1\neq \bzero$, and for every $k>1$, the vector $\ba_k$ does not belong to the span of $\ba_1, \ba_2, \cdots, \ba_{k-1}$.
\end{definition}

In the study of linear algebra, every vector space has a basis and every vector is a linear combination of members of the basis. We then define span and dimension of subspace via the basis.

\begin{definition}[Span]
	If every vector $\bv$ in subspace $\mathcal{V}$ can be expressed as a linear combination of $\{\ba_1, \ba_2, \cdots, \ba_k\}$, then $\{\ba_1, \ba_2, \cdots, \ba_k\}$ is said to span $\mathcal{V}$.
\end{definition}

\begin{definition}[Basis and Dimension]
A set of vectors $\{\ba_1, \ba_2, \cdots, \ba_k\}$ is called a basis of $\mathcal{V}$ if they are linearly independent and span $\mathcal{V}$. 
Every basis of a given subspace has the same number of vectors, and the number of vectors in any basis is called the dimension of the subspace $\mathcal{V}$. By convention, the subspace $\{\bzero\}$ is said to  have dimension zero. Furthermore, every subspace of nonzero dimension has a basis that is orthogonal, i.e., the basis of a subspace can be chosen orthogonal.
\end{definition}

\begin{definition}[Column Space (Range)]
	If $\bA$ is an $n \times p$ real matrix, we define the column space (or range) of $\bA$ to be the set spanned by its columns:
	\begin{equation*}
		\mathcal{C} (\bA) = \{ \by\in \mathbb{R}^n: \exists \bx \in \mathbb{R}^p, \, \by = \bA \bx \}.
	\end{equation*}
And the row space of $\bA$ is the set spanned by its rows, which is equal to the column space of $\bA^\top$:
	\begin{equation*}
	\mathcal{C} (\bA^\top) = \{ \bx\in \mathbb{R}^p: \exists \by \in \mathbb{R}^n, \, \bx = \bA^\top \by \}.
\end{equation*}
\end{definition}

\begin{definition}[Null Space (Kernel)]
	If $\bA$ is an $n \times p$ real matrix, we define the null space (or kernel) of $\bA$ to be the set:
	\begin{equation*}
		\nspace (\bA) = \{\by \in \mathbb{R}^p:  \, \bA \by = \bzero \}.
	\end{equation*}
And the null space of $\bA^\top$ is defined as 
	\begin{equation*}
	\nspace (\bA^\top) = \{\bx \in \mathbb{R}^n:  \, \bA^\top \bx = \bzero \}.
\end{equation*}
\end{definition}

Both the column space of $\bA$ and the null space of $\bA^\top$ are subspaces of $\real^p$. In fact, every vector in $\nspace(\bA^\top)$ is perpendicular to $\cspace(\bA)$ and vice versa.\footnote{Every vector in $\nspace(\bA)$ is also perpendicular to $\cspace(\bA^\top)$ and vice versa.}

\begin{definition}[Rank]
	The $rank$ of a matrix $\bA\in \real^{n\times p}$ is the dimension of the column space of $\bA$. The rank of $\bA$ is equal to the maximal number of linearly independent columns of $\bA$, and is also the maximal number of linearly independent rows of $\bA$. The matrix $\bA$ and its transpose $\bA^\top$ have the same rank. We say that $\bA$ has full rank, if its rank is equal to $min\{n,p\}$. In another word, this is true if and only if either all the columns of $\bA$ are linearly independent, or all the rows of $\bA$ are linearly independent. Specifically, given a vector $\bu \in \real^n$ and a vector $\bv \in \real^p$, then the $n\times p$ matrix $\bu\bv^\top$ is of rank 1.
\end{definition}

\begin{definition}[Orthogonal Complement in General]
The orthogonal complement $\mathcalV^\perp$ of a subspace $\mathcalV$ contains every vector that is perpendicular to $\mathcalV$. 
That is,
$$
\mathcalV^\perp = \{\bv | \bv^\top\bu=0, \,\,\, \forall \bu\in \mathcalV  \}.
$$
The two subspaces are disjoint that span  the entire space. The dimensions of $\mathcalV$ and $\mathcalV^\perp$ add to the dimension of the whole space. Furthermore, $(\mathcalV^\perp)^\perp=\mathcalV$.
\end{definition}

\begin{definition}[Orthogonal Complement of Column Space]
	If $\bA$ is an $n \times p$ real matrix, the orthogonal complement of $\mathcal{C}(\bA)$, $\mathcal{C}^{\bot}(\bA)$ is the subspace defined as:
	\begin{equation*}
		\begin{aligned}
			\mathcal{C}^{\bot}(\bA) &= \{\by\in \mathbb{R}^n: \, \by^\top \bA \bx=\bzero, \, \forall \bx \in \mathbb{R}^p \} \\
			&=\{\by\in \mathbb{R}^n: \, \by^\top \bv = \bzero, \, \forall \bv \in \mathcal{C}(\bA) \}.
		\end{aligned}
	\end{equation*}
\end{definition}
%
%
%


\begin{definition}[Orthogonal Matrix]
	A real square matrix $\bQ$ is an orthogonal matrix if the inverse of $\bQ$ equals to the transpose of $\bQ$, that is $\bQ^{-1}=\bQ^\top$ and $\bQ\bQ^\top = \bQ^\top\bQ = \bI$. In another word, suppose $\bQ=[\bq_1, \bq_2, \cdots, \bq_n]$ where $\bq_i \in \real^n$ for all $i \in \{1, 2, \cdots, n\}$, then $\bq_i^\top \bq_j = \delta(i,j)$ with $\delta(i,j)$ being the Kronecker delta function. If $\bQ$ contains only $\gamma$ of these columns with $\gamma<n$, then $\bQ^\top\bQ = \bI_\gamma$ stills holds with $\bI_\gamma$ being the $\gamma\times \gamma$ identity matrix. But $\bQ\bQ^\top=\bI$ will not be true.
\end{definition}

\section{Fundamental Theorem of Linear Algebra}

\subsection{Dimension of Column Space and Row Space}\label{append:row-equal-column}
\label{app:theorem}

In this section, we prove Lemma~\ref{lemma:equal-dimension-rank} that the dimension of the column space of a matrix $\bX\in \real^{n\times p}$ is equal to the dimension of its row space, i.e., the row rank and the column rank of a matrix $\bX$ are equal.
\begin{svgraybox}
	\begin{lemma}[Dimension of Column Space and Row Space]\label{lemma:equal-dimension-rank}
		The dimension of the column space of a matrix $\bX\in \real^{n\times p}$ is equal to the dimension of its
		row space, i.e., the row rank and the column rank of a matrix $\bX$ are equal.
	\end{lemma}
\end{svgraybox}
\begin{proof}[of Lemma~\ref{lemma:equal-dimension-rank}]
	We first notice that the null space of $\bX$ is orthogonal complementary to the row space of $\bX$: $\nspace(\bX) \bot \cspace(\bX^\top)$ (where the row space of $\bX$ is exactly the column space of $\bX^\top$), that is, vectors in the null space of $\bX$ are orthogonal to vectors in the row space of $\bX$. To see this, suppose $\bX$ has rows $\ba_1^\top, \ba_2^\top, \cdots, \ba_n^\top$ and $\bX=[\ba_1^\top; \ba_2^\top; \cdots; \ba_n^\top]$. For any vector $\bbeta\in \nspace(\bX)$, we have $\bX\bbeta = \bzero$, that is, $[\ba_1^\top\bbeta; \ba_2^\top\bbeta; \cdots; \ba_n^\top\bbeta]=\bzero$. And since the row space of $\bX$ is spanned by $\ba_1^\top, \ba_2^\top, \cdots, \ba_n^\top$. Then $\bbeta$ is perpendicular to any vectors from $\cspace(\bX^\top)$ which means $\nspace(\bX) \bot \cspace(\bX^\top)$.
	
	Now suppose, the dimension of the row space of $\bX$ is $r$. \textcolor{blue}{Let $\br_1, \br_2, \cdots, \br_r$ be a set of vectors in $\real^n$ and form a basis for the row space}. Then the $r$ vectors $\bX\br_1, \bX\br_2, \cdots, \bX\br_r$ are in the column space of $\bX$, furthermore, they are linearly independent. To see this, suppose we have a linear combination of the $r$ vectors: $\beta_1\bA\br_1 + \beta_2\bA\br_2+ \cdots+ \beta_r\bA\br_r=0$, that is, $\bX(\beta_1\br_1 + \beta_2\br_2+ \cdots+ \beta_r\br_r)=0$ and the vector $\bv=\beta_1\br_1 + \beta_2\br_2+ \cdots+ \beta_r\br_r$ is in the null space of $\bX$. But since $\{\br_1, \br_2, \cdots, \br_r\}$ is a basis for the row space of $\bX$, $\bv$ is thus also in the row space of $\bX$. We have shown that vectors from the null space of $\bX$ is perpendicular to vectors from row space of $\bX$, thus $\bv^\top\bv=0$ and $\beta_1=\beta_2=\cdots=\beta_r=0$. Then \textcolor{blue}{$\bX\br_1, \bX\br_2, \cdots, \bX\br_r$ are in the column space of $\bX$ and they are independent} which means column space of $\bX$ is larger than $r$. This result shows that \textbf{row rank of $\bX\leq $ column rank of $\bX$}. 
	
	If we apply this process again for $\bX^\top$. We will have \textbf{column rank of $\bX\leq $ row rank of $\bX$}. We complete the proof. 
\end{proof}

A further information can be drawn from this proof is that if $\{\br_1, \br_2, \cdots, \br_r\}$ is a set of vectors in $\real^p$ that forms a basis for the row space, then \textcolor{blue}{$\{\bX\br_1, \bX\br_2, \cdots, \bX\br_r\}$ is a basis for the column space of $\bX$}. We formulate this finding into the following lemma.

\begin{svgraybox}
	\begin{lemma}[Column Basis from Row Basis]\label{lemma:column-basis-from-row-basis}
		For any matrix $\bX\in \real^{n\times p}$, let $\{\br_1, \br_2, \cdots, \br_r\}$ be a set of vectors in $\real^p$ which forms a basis for the row space, then $\{\bX\br_1, \bX\br_2, \cdots, \bX\br_r\}$ is a basis for the column space of $\bX$.
	\end{lemma}
\end{svgraybox}

\subsection{Fundamental Theorem of Linear Algebra}\label{appendix:fundamental-rank-nullity}
\begin{figure}[h!]
	\centering
	\includegraphics[width=0.98\textwidth]{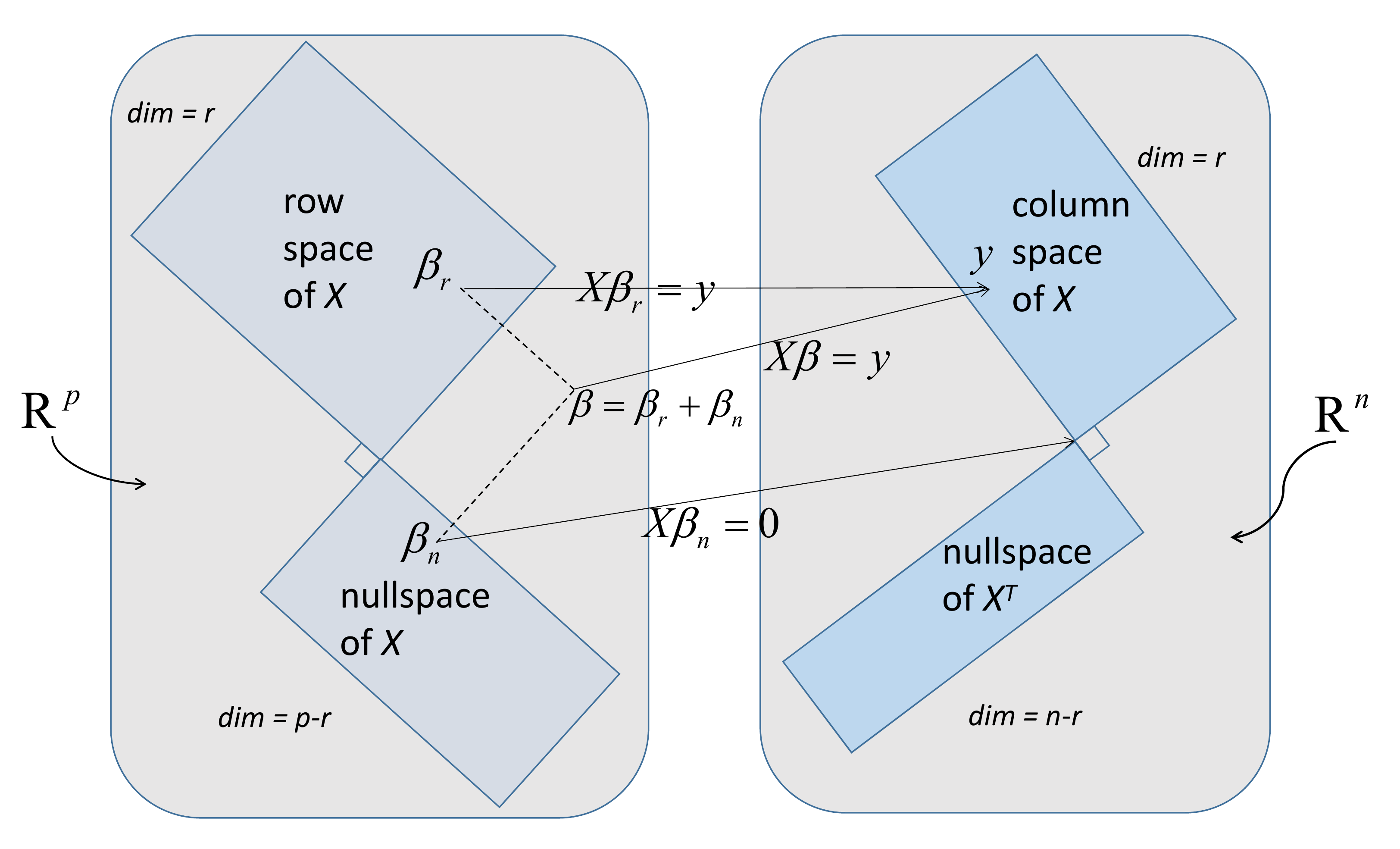}
	\caption{\textbf{First Figure}: Two pairs of orthogonal subspaces in $\real^p$ and $\real^n$. $\dim(\cspace(\bX^\top)) + \dim(\nspace(\bX)) = p$ and $dim(\nspace(\bX^\top))+dim(\cspace(\bX))=n$. The null space component goes to zero as $\bX\bbeta_{\bn} = \bzero \in \real^n$. The row space component goes to the column space as $\bX\bbeta_{\br} = \bX(\bbeta_{\br} + \bbeta_{\bn})=\by\in \cspace(\bX)$.}
	\label{fig:lafundamental-ls}
\end{figure}

As a recap, for any matrix $\bX\in \real^{n\times p}$, it can be easily verified that any vector in the row space of $\bX$ is perpendicular to any vector in the null space of $\bX$. Suppose $\bbeta_n \in \nspace(\bX)$, then $\bX\bbeta_n = \bzero$ such that $\bbeta_n$ is perpendicular to every row of $\bX$ which agrees with our claim. 

Similarly, we can also show that any vector in the column space of $\bX$ is perpendicular to any vector in the null space of $\bX^\top$. Furthermore, the column space of $\bX$ and the null space of $\bX^\top$ span the whole $\real^n$ space which is known as the fundamental theorem of linear algebra.

The fundamental theorem contains two parts, the dimension of the subspaces and the orthogonality of the subspaces. The orthogonality can be easily verified as we have shown in the beginning of this section. When the row space has dimension $r$, the null space has dimension $p-r$. This cannot be easily stated and the proof of the above theorem is provided as follows. 
\begin{svgraybox}
	\begin{theorem}[Fundamental Theorem of Linear Algebra]\label{theorem:fundamental-linear-algebra}
		
		Orthogonal Complement and Rank-Nullity Theorem: for any matrix $\bX\in \real^{n\times p}$, we have 
		
		$\bullet$ $\nspace(\bX)$ is orthogonal complement to the row space $\cspace(\bX^\top)$ in $\real^p$: $dim(\nspace(\bX))+dim(\cspace(\bX^\top))=p$;
		
		$\bullet$ $\nspace(\bX^\top)$ is orthogonal complement to the column space $\cspace(\bX)$ in $\real^n$: $dim(\nspace(\bX^\top))+dim(\cspace(\bX))=n$;
		
		$\bullet$ For rank-$r$ matrix $\bX$, $dim(\cspace(\bX^\top)) = dim(\cspace(\bX)) = r$, that is, $dim(\nspace(\bX)) = p-r$ and $dim(\nspace(\bX^\top))=n-r$.
	\end{theorem}
\end{svgraybox}

\begin{proof}[of Theorem~\ref{theorem:fundamental-linear-algebra}]
	Follow from the proof of Lemma~\ref{lemma:equal-dimension-rank} in Section~\ref{append:row-equal-column}. Let $\{\br_1, \br_2, \cdots, \br_r\}$ be a set of vectors in $\real^p$ that form a basis for the row space, then \textcolor{blue}{$\{\bX\br_1, \bX\br_2, \cdots, \bX\br_r\}$ is a basis for the column space of $\bX$}. Let $\bn_1, \bn_2, \cdots, \bn_k \in \real^p$ form a basis for the null space of $\bX$. Follow again from the proof of Lemma~\ref{lemma:equal-dimension-rank}, $\nspace(\bX) \bot \cspace(\bX^\top)$, thus, $\br_1, \br_2, \cdots, \br_r$ are perpendicular to $\bn_1, \bn_2, \cdots, \bn_k$. Then, $\{\br_1, \br_2, \cdots, \br_r, \bn_1, \bn_2, \cdots, \bn_k\}$ is linearly independent in $\real^p$.
	
	For any vector $\bbeta\in \real^p $, $\bX\bbeta$ is in the column space of $\bX$. Then it can be expressed by a combination of $\bX\br_1, \bX\br_2, \cdots, \bX\br_r$: $\bX\bbeta = \sum_{i=1}^{r}a_i\bX\br_i$ which states that $\bX(\bbeta-\sum_{i=1}^{r}a_i\br_i) = \bzero$ and $\bbeta-\sum_{i=1}^{r}a_i\br_i$ is thus in $\nspace(\bX)$. Since $\bn_1, \bn_2, \cdots, \bn_k$ is a basis for the null space of $\bX$, $\bbeta-\sum_{i=1}^{r}a_i\br_i$ can be expressed by a combination of $\bn_1, \bn_2, \cdots, \bn_k$: $\bbeta-\sum_{i=1}^{r}a_i\br_i = \sum_{j=1}^{k}b_j \bn_j$, i.e., $\bbeta=\sum_{i=1}^{r}a_i\br_i + \sum_{j=1}^{k}b_j \bn_j$. That is, any vector $\bbeta\in \real^p$ can be expressed by $\{\br_1, \br_2, \cdots, \br_r, \bn_1, \bn_2, \cdots, \bn_k\}$ and the set forms a basis for $\real^p$. Thus the dimension sum to $p$: $r+k=p$, i.e., $dim(\nspace(\bX))+dim(\cspace(\bX^\top))=p$. Similarly, we can prove $dim(\nspace(\bX^\top))+dim(\cspace(\bX))=n$.
\end{proof}

Figure~\ref{fig:lafundamental-ls} demonstrates two pairs of such orthogonal subspaces and shows how $\bX$ takes $\bbeta$ into the column space. The dimensions of the row space of $\bX$ and the null space of $\bX$ add to $p$. And the dimensions of the column space of $\bX$ and the null space of $\bX^\top$ add to $n$. The null space component goes to zero as $\bX\bbeta_{\bn} = \bzero \in \real^n$ which is the intersection of the column space of $\bX$ and the null space of $\bX^\top$. The row space component goes to column space as $\bX\bbeta_{\br} = \bX(\bbeta_{\br} + \bbeta_{\bn})=\by\in \cspace(\bX)$.

\section{SVD in the Fundamental Theorem of Linear Algebra}\label{section:svd-funda}
By QR decomposition, we factor matrix into an orthogonal matrix (see \citep{strang1993introduction, trefethen1997numerical, strang2021every, lu2021numerical, lu2021rigorous}, and the complexity and applications of QR decomposition). Instead of factoring matrix into one orthogonal matrix, SVD gives rise to two orthogonal matrices. We illustrate the result of SVD in the following theorem. And in Appendix~\ref{appendix:SVD}, we provide a rigorous proof for the existence of SVD.

\begin{figure}[h!]
	\centering  
	\vspace{-0.35cm} 
	\subfigtopskip=2pt 
	\subfigbottomskip=2pt 
	\subfigcapskip=-5pt 
	\subfigure[Reduced SVD decomposition]{\label{fig:svdhalf}
		\includegraphics[width=0.47\linewidth]{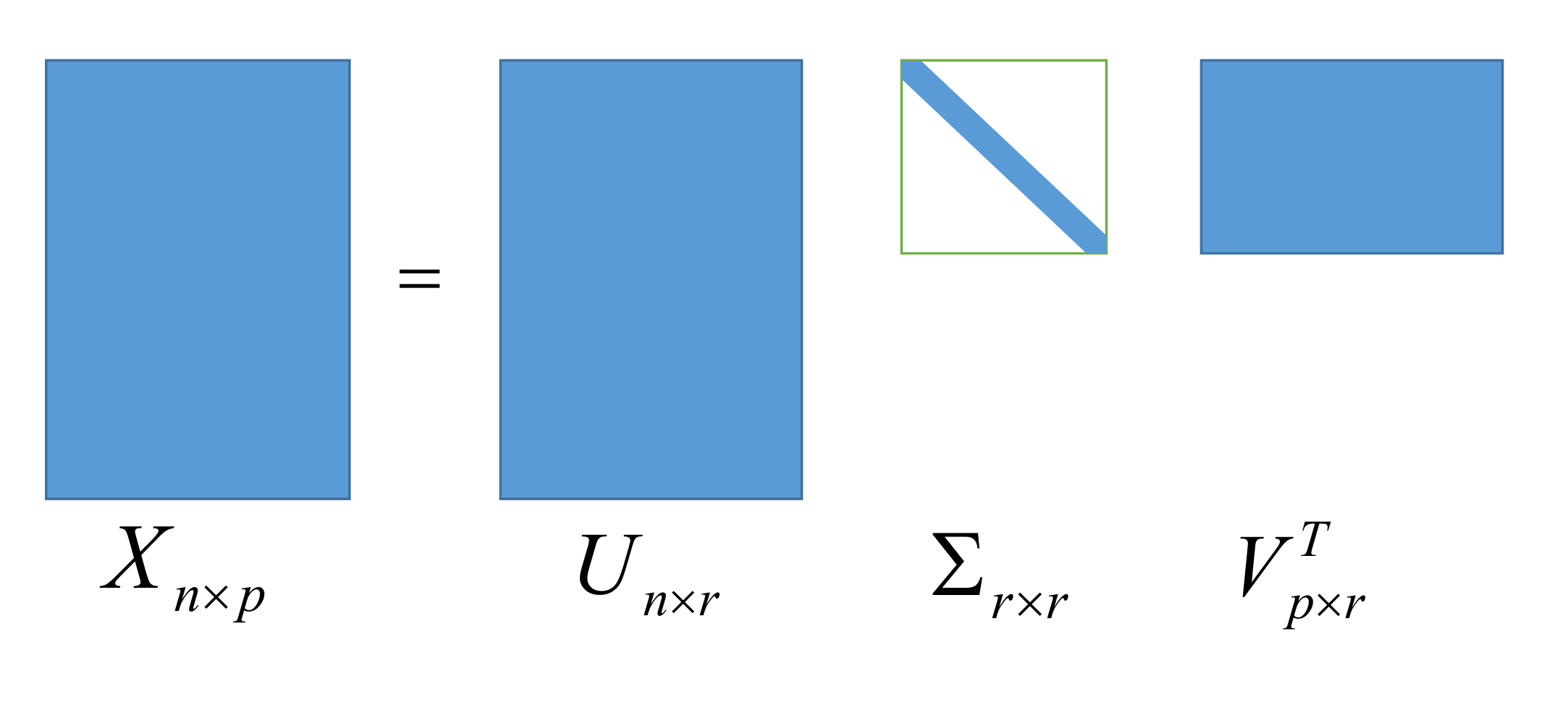}}
	\quad 
	\subfigure[Full SVD decomposition]{\label{fig:svdall}
		\includegraphics[width=0.47\linewidth]{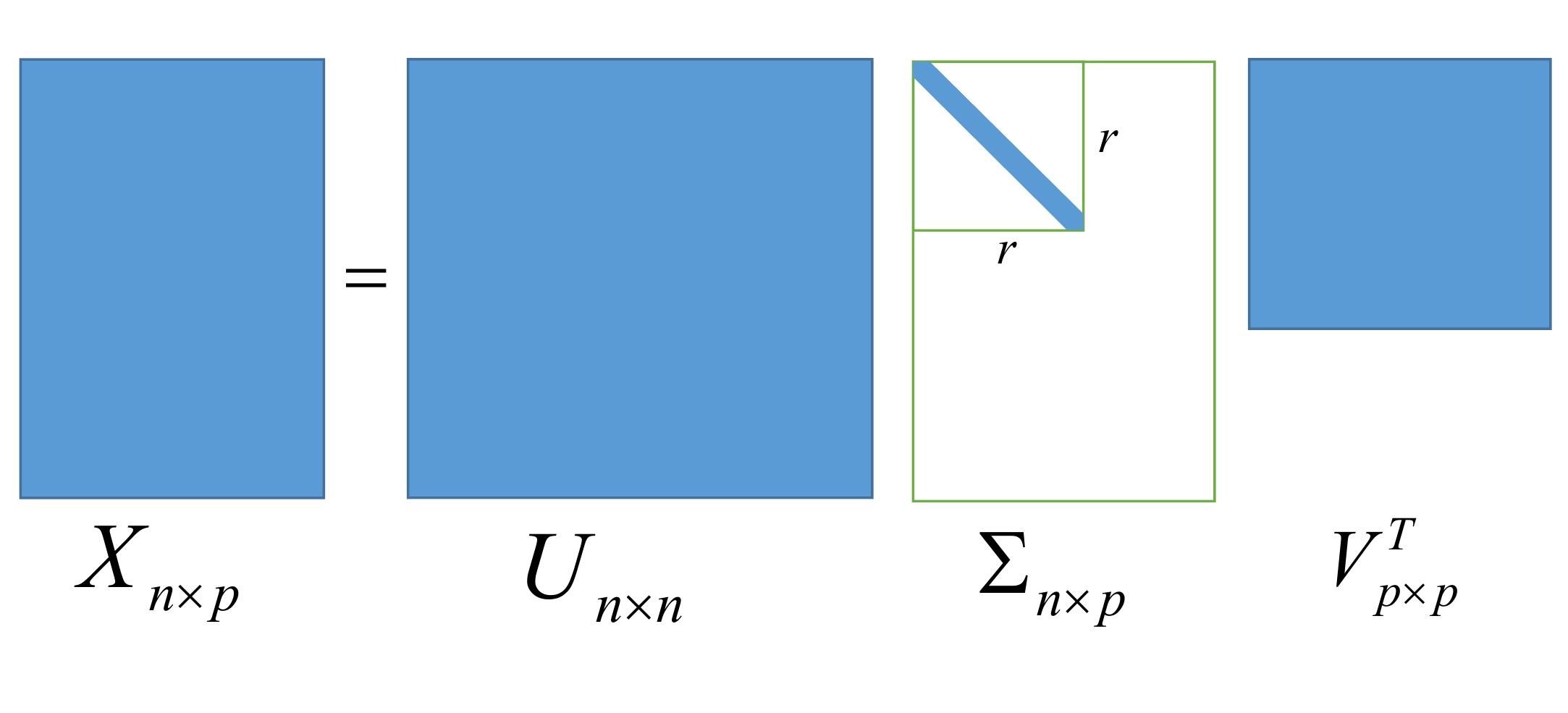}}
	\caption{Comparison of reduced and full SVD. White entries are zero and blue entries are not necessarily zero. In reduced SVD, we remove $n-r$ silent columns from $\bU$ and $p-r$ silent columns from $\bV$.}
	\label{fig:svd-comparison}
\end{figure}

\begin{svgraybox}
	\begin{theorem}[Full SVD for Rectangular Matrices]\label{theorem:full_svd_rectangular}
		For every real $n\times p$ matrix $\bX$ with rank $r$, then matrix $\bX$ can be factored as
		$$
		\bX = \bU \bSigma \bV^\top,
		$$ 
		where the left-upper side of $\bSigma\in $\textcolor{blue}{$\real^{n\times p}$} is a diagonal matrix, that is $\bSigma=\begin{bmatrix}
			\bSigma_1 & \bzero \\
			\bzero & \bzero
		\end{bmatrix}$ where $\bSigma_1=diag(\sigma_1, \sigma_2 \cdots, \sigma_r)\in \real^{r\times r}$ with $\sigma_1 \geq \sigma_2 \geq \cdots \geq \sigma_r$ and 
		
		$\bullet$ $\sigma_i$'s are the nonzero \textbf{singular values} of matrix $\bX$, in the meantime, they are the (positive) square roots of the nonzero \textbf{eigenvalues} of $\trans{\bX} \bX$ and $ \bX \trans{\bX}$. 
		
		$\bullet$ $\bU\in \textcolor{blue}{\real^{n\times n}}$ contains the $r$ eigenvectors of $\bX\bX^\top$ corresponding to the $r$ nonzero eigenvalues of $\bX\bX^\top$ \textcolor{blue}{and $n-r$ extra orthonormal vectors from $\nspace(\bX^\top)$}. 
		
		$\bullet$ $\bV\in \textcolor{blue}{\real^{p\times p}}$ contains the $r$ eigenvectors of $\bX^\top\bX$ corresponding to the $r$ nonzero eigenvalues of $\bX^\top\bX$ \textcolor{blue}{and $p-r$ extra orthonormal vectors from $\nspace(\bX)$}.
		
		$\bullet$ Moreover, the columns of $\bU$ and $\bV$ are called the \textbf{left and right singular vectors} of $\bX$, respectively. 
		
		$\bullet$ Further, the columns of $\bU$ and $\bV$ are orthonormal, and \textcolor{blue}{$\bU$ and $\bV$ are orthogonal matrices}. 
		
		In particular, we can write the matrix decomposition by $ \bX = \bU \bSigma \bV^\top = \sum_{i=1}^r \sigma_i \bu_i \bv_i^\top$, which is a sum of $r$ rank-one matrices.
	\end{theorem}
\end{svgraybox}

\begin{figure}[h!]
	\centering
	\includegraphics[width=0.98\textwidth]{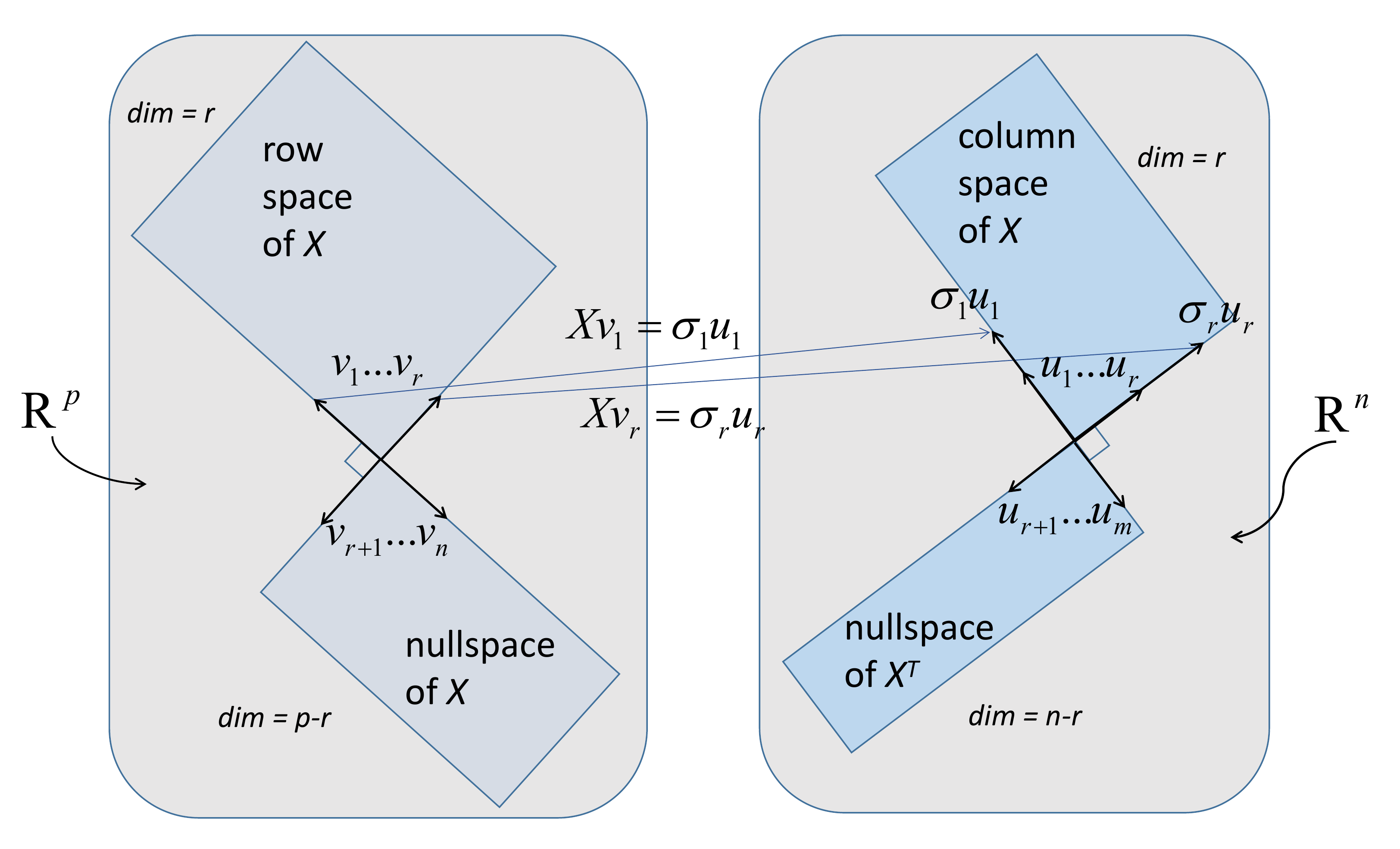}
	\caption{\textbf{Second Figure}: Orthonormal bases that diagonalize $\bX$ from SVD. $\{\bv_1, \bv_2, \cdots, \bv_r\} $ is an orthonormal basis of $\cspace(\bX^\top)$, and $\{\bu_1,\bu_2, \cdots,\bu_r\}$ is an orthonormal basis of $\cspace(\bX)$. $\bX$ transfers the row basis $\bv_i$ into column basis $\bu_i$ by $\sigma_i\bu_i=\bX\bv_i$ for all $i\in \{1, 2, \cdots, r\}$.}
	\label{fig:lafundamental3-SVD}
\end{figure}

There is a version of reduced SVD (see Theorem~\ref{theorem:reduced_svd_rectangular} in Appendix~\ref{appendix:SVD}) by removing the silent columns in $\bU$ and $\bV$.  
The comparison of reduced and full SVD is shown in Figure~\ref{fig:svd-comparison} where white entries are zero and blue entries are not necessarily zero.

The fundamental theorem of linear algebra tells us there exists a basis for each subspace.
Specifically, from SVD, we can find an orthonormal basis for each subspace.

\begin{svgraybox}
	\begin{lemma}[Four Orthonormal Basis in SVD]\label{lemma:svd-four-orthonormal-Basis}
		Given the full SVD of matrix $\bX = \bU \bSigma \bV^\top$, where $\bU=[\bu_1, \bu_2, \cdots,\bu_n]$ and $\bV=[\bv_1, \bv_2, \cdots, \bv_p]$ are the column partitions of $\bU$ and $\bV$. Then, we have the following property:
		
		$\bullet$ $\{\bv_1, \bv_2, \cdots, \bv_r\} $ is an orthonormal basis of the row space,  $\cspace(\bX^\top)$;
		
		$\bullet$ $\{\bv_{r+1},\bv_{r+2}, \cdots, \bv_p\}$ is an orthonormal basis of $\nspace(\bX)$;
		
		$\bullet$ $\{\bu_1,\bu_2, \cdots,\bu_r\}$ is an orthonormal basis of the column space, $\cspace(\bX)$;
		
		$\bullet$ $\{\bu_{r+1}, \bu_{r+2},\cdots,\bu_n\}$ is an orthonormal basis of $\nspace(\bX^\top)$. 
	\end{lemma}
\end{svgraybox}

%
%
%

\begin{proof}[of Lemma~\ref{lemma:svd-four-orthonormal-Basis}]
	From Lemma~\ref{lemma:rank-of-symmetric}, for symmetric matrix $\bX^\top\bX$, $\cspace(\bX^\top\bX)$ is spanned by the eigenvectors, thus $\{\bv_1,\bv_2 \cdots, \bv_r\}$ is an orthonormal basis of $\cspace(\bX^\top\bX)$.
	
	Since, 
	
	1. $\bX^\top\bX$ is symmetric, then the row space of $\bX^\top\bX$ is equal to the column space of $\bX^\top\bX$. 
	
	2. All rows of $\bX^\top\bX$ are combination of rows of $\bX$, so the row space of $\bX^\top\bX$ $\subseteq$ row space of $\bX$, i.e., $\cspace(\bX^\top\bX) \subseteq \cspace(\bX^\top)$. 
	
	3. Since $rank(\bX^\top\bX) = rank(\bX)$ by Lemma~\ref{lemma:rank-of-ata-x}, we then have 
	
	The row space of $\bX^\top\bX$ = the column space of $\bX^\top\bX$ =  the row space of $\bX$, i.e., $\cspace(\bX^\top\bX) = \cspace(\bX^\top)$. Thus $\{\bv_1, \bv_2,\cdots, \bv_r\}$ is an orthonormal basis of $\cspace(\bX^\top)$. 
	
	Further, the space spanned by $\{\bv_{r+1}, \bv_{r+2},\cdots, \bv_n\}$ is an orthogonal complement to the space spanned by $\{\bv_1,\bv_2, \cdots, \bv_r\}$, so $\{\bv_{r+1},\bv_{r+2}, \cdots, \bv_n\}$ is an orthonormal basis of $\nspace(\bX)$. 
	
	If we apply this process to $\bX\bX^\top$, we will prove the rest claims in the lemma. As an alternative way, we realize that $\{\bu_1,\bu_2, \cdots,\bu_r\}$ is a basis for the column space of $\bX$ by Lemma~\ref{lemma:column-basis-from-row-basis} \footnote{As a recap, for any matrix $\bX$, let $\{\br_1, \br_2, \cdots, \br_r\}$ be a set of vectors in $\real^p$ which forms a basis for the row space, then $\{\bX\br_1, \bX\br_2, \cdots, \bX\br_r\}$ is a basis for the column space of $\bX$.}, since $\bu_i = \frac{\bX\bv_i}{\sigma_i},\, \forall i \in\{1, 2, \cdots, r\}$. 
\end{proof}

The relationship of the four subspaces is demonstrated in Figure~\ref{fig:lafundamental3-SVD} where $\bX$ transfers the row basis $\bv_i$ into column basis $\bu_i$ by $\sigma_i\bu_i=\bX\bv_i$ for all $i\in \{1, 2, \cdots, r\}$.

\section{Least Squares in the Fundamental Theorem of Linear Algebra}\label{section:ls-fundation-theorem}
\begin{figure}[h!]
	\centering
	\includegraphics[width=0.98\textwidth]{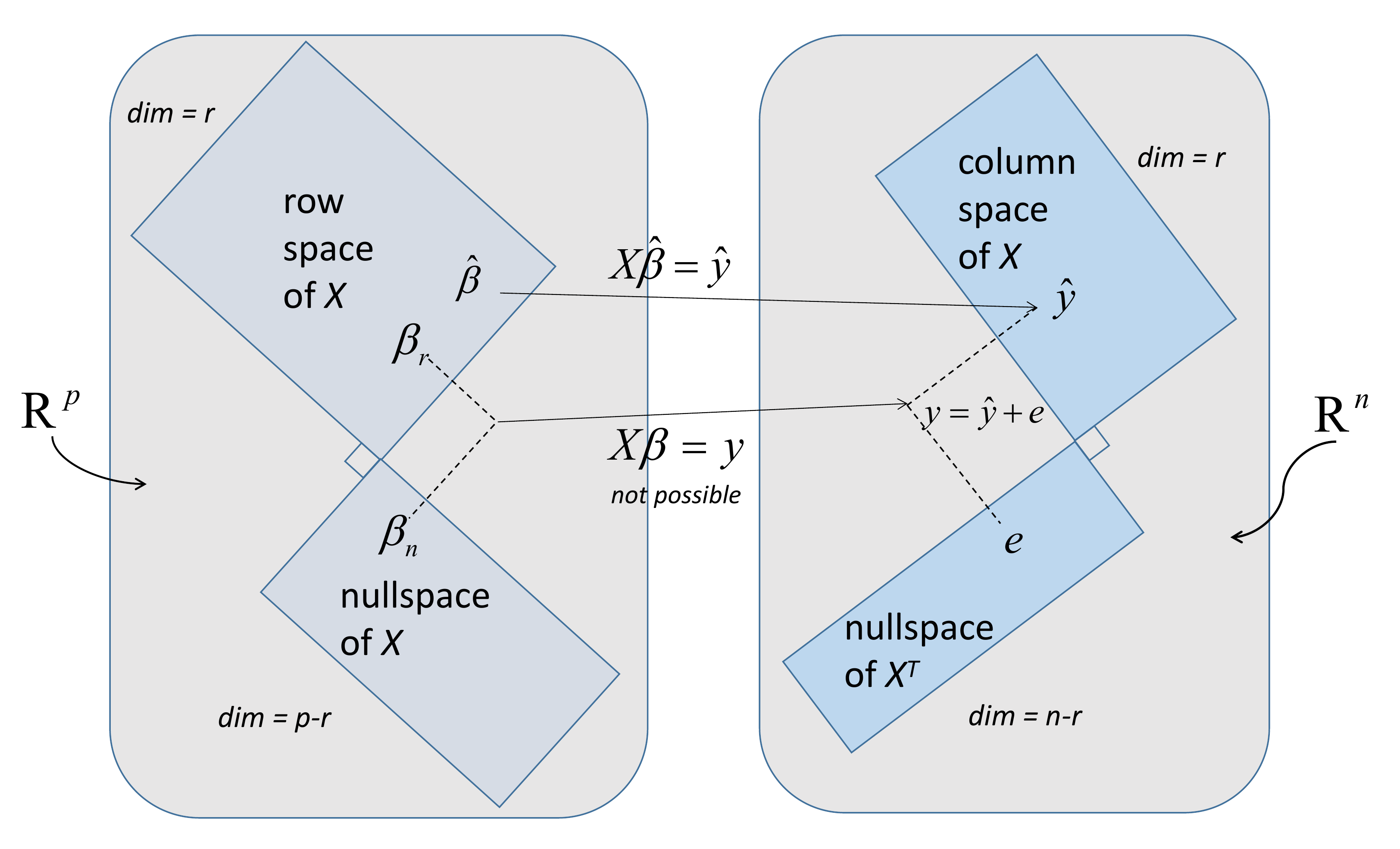}
	\caption{\textbf{Third Figure}: Least Squares, row space to column space view. $\hat{\bbeta}$ minimizes $||\by-\bX\bbeta||^2$ such that $\hat{\bbeta}$ is in the row space of $\bX$.}
	\label{fig:lafundamental2-LS}
\end{figure}

Let's consider the overdetermined system $\by = \bX\bbeta $, where $\bX\in \real^{n\times p}$ is the input data matrix, $\by\in \real^n$ is the observation matrix (target matrix), and the sample number $n$ is larger than the dimension number $p$. $\bbeta$ is a vector of weights of the linear model. Normally $\bX$ will have full column rank since the data from real world has large chance to be unrelated. The solution of least squares is to make the error $\by-\bX\bbeta$ as small as possible with respect to the mean squared error. $\bX\bbeta$ is a combination of the columns of $\bX$, as a result, $\bX\bbeta$ can never leave the column space of $\bX$ such that we should choose the closest point to $\by$ in the column space \citep{strang1993fundamental, lu2021rigorous}. This point is the projection $\hat{\by}$ of $\by$ onto the column space of $\bX$. Then the error vector $\be=\by-\hat{\by}$ has minimal length. In another word, the best combination $\hat{\by} = \bX\hat{\bbeta}$ is the projection of $\by$ onto the column space. The error $\be$ is perpendicular to the column space. Therefore \textcolor{blue}{$\be=\by-\bX\hat{\bbeta}$ is in the null space of $\bX^\top$}:
$$
\bX^\top(\by-\bX\hat{\bbeta}) = \bzero  \qquad \text{or} \qquad \bX^\top\by=\bX^\top\bX\hat{\bbeta},
$$
which is also known as \textbf{the normal equation}. The ordinary least squares (OLS) estimator $\hat{\bbeta}$ from this normal equation is $\hat{\bbeta} = (\bX^\top\bX)^{-1}\bX^\top\by$. The relationship between $\be$ and $\hat{\by}$ is shown in Figure~\ref{fig:lafundamental2-LS} where $\by$ is split into $\hat{\by}+\be$. In addition, since the column space of $\bX$ and the null space of $\bX^\top$ span the whole $\real^n$ space, any vector $\by$ can be split into a vector $\hat{\by}$ in the column space of $\bX$ and a vector $\be$ in the null space of $\bX^\top$. Moreover, it can be shown that the OLS estimator $\hat{\bbeta}$ is in the row space of $\bX$, i.e., it cannot be split into a combination of two components that are in the row space of $\bX$ and the null space of $\bX$ respectively (see $\hat{\bbeta}$ via the pseudo-inverse of $\bX$ in Section~\ref{section:ls-via-svd}, $\hat{\bbeta}$ is a linear combination of the orthonormal basis of the row space).

To conclude, we avoid solving the equation $\by = \bX\bbeta$ by removing $\be$ from $\by$ and solve $\hat{\by} = \bX\hat{\bbeta}$ instead:
\begin{equation}
	\bX\bbeta=\by = \hat{\by}+\be \,\, \mathrm{is\, impossible;} \qquad \bX\hat{\bbeta}=\hat{\by}\,\, \mathrm{is\, possible.} \nonumber
\end{equation}


%
%
%
%
%

\section{Least Squares via the Full SVD for Rank Deficient Matrices}\label{section:ls-via-svd}
In the previous section, we assume $\bX$ has full rank. However, if $\bX$ does not have full column rank, $\bX^\top\bX$ is not invertible. We thus can use SVD decomposition of $\bX$ to solve the least squares problem with rank-deficient $\bX$. And we illustrate this rank-deficient least squares method in the following theorem. 

\begin{svgraybox}
\begin{theorem}[LS via SVD for Rank Deficient Matrix]\label{theorem:svd-deficient-rank}
Let $\bX\in \real^{n\times p}$, $\bX=\bU\bSigma\bV^\top$ is its full SVD decomposition with $\bU\in\real^{n\times n}$ and $\bV\in \real^{p\times p}$ being orthogonal matrices and $rank(\bX)=r \leq \min\{n,p\}$. Suppose $\bU=[\bu_1, \bu_2, \cdots, \bu_n]$, $\bV=[\bv_1, \bv_2, \cdots, \bv_p]$ and $\by\in \real^n$, then the LS solution to $\bX\bbeta=\by$ is given by 
$$
\hat{\bbeta} = \sum_{i=1}^{r}\frac{\bu_i^\top \by}{\sigma_i}\bv_i = \bV\bSigma^+\bU^\top \by, 
$$
where the upper-left side of $\bSigma^+ \in \real^{p\times n}$ is a diagonal matrix $diag(\frac{1}{\sigma_1}, \frac{1}{\sigma_2}, \cdots, \frac{1}{\sigma_r})$.
\end{theorem}
\end{svgraybox}

\begin{proof}[of Theorem~\ref{theorem:svd-deficient-rank}]
	Write out the loss to be minimized
	$$
	\begin{aligned}
		||\bX\bbeta-\by||^2 &= (\bX\bbeta-\by)^\top(\bX\bbeta-\by)\\
		&=(\bX\bbeta-\by)^\top\bU\bU^\top (\bX\bbeta-\by) \qquad &(\text{Since $\bU$ is an orthogonal matrix})\\
		&=||\bU^\top \bX \bbeta-\bU^\top\by||^2 \qquad &(\text{Invariant under orthogonal})\\
		&=||\bU^\top \bX \bV\bV^\top \bbeta-\bU^\top\by||^2 \qquad &(\text{Since $\bV$ is an orthogonal matrix})\\
		&=||\bSigma\balpha - \bU^\top\by||^2  \qquad &(\text{Let $\balpha=\bV^\top \bbeta$})\\
		&=\sum_{i=1}^{r}(\sigma_i\balpha_i - \bu_i^\top\by)^2 +\sum_{i=r+1}^{n}(\bu_i^\top \by)^2. \qquad &(\text{Since $\sigma_{r+1}=\sigma_{r+2}= \cdots= \sigma_n=0$})
	\end{aligned}
	$$
	Since $\bbeta$ only appears in $\balpha$, we just need to set $\balpha_i = \frac{\bu_i^\top\by}{\sigma_i}$ for all $i\in \{1, 2, \cdots, r\}$ to minimize the above equation. For any value of $\balpha_{r+1}, \balpha_{r+2}, \cdots, \balpha_{p}$, it won't change the result. From the regularization point of view we can set them to be 0 (to get minimal length). This gives us the LS solution via SVD:
	$$
	\hat{\bbeta} = \sum_{i=1}^{r}\frac{\bu_i^\top \by}{\sigma_i}\bv_i=\bV\bSigma^+\bU^\top \by = \bX^+\by,
	$$
	where $\bX^+ = \bV\bSigma^+\bU^\top\in \real^{p\times n}$ is known as the \textbf{pseudo-inverse} of $\bX$.  Please refer to Section~\ref{appendix:pseudo-inverse} for a detailed discussion about pseudo-inverse where we also prove that \textbf{the column space of $\bX^+$ is equal to the row space of $\bX$, and the row space of $\bX^+$ is equal to the column space of $\bX$}.
\end{proof}

\begin{figure}[h!]
	\centering
	\includegraphics[width=0.98\textwidth]{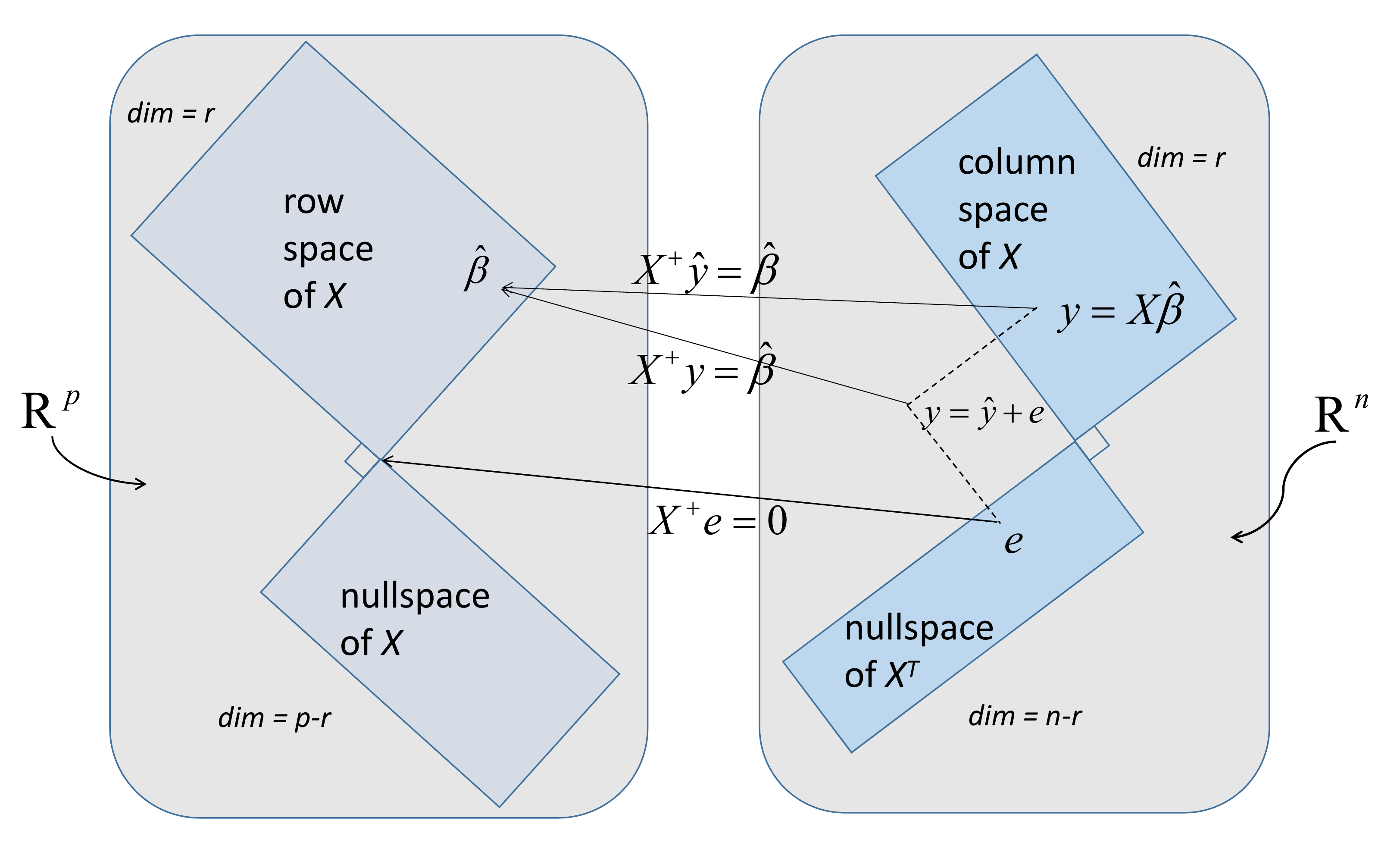}
	\caption{\textbf{Fourth Figure}: Least Squares, column space to row space view. $\bX^+$: Pseudo-inverse of $\bX$.}
	\label{fig:lafundamental4-LS-SVD}
\end{figure}

\begin{svgraybox}
\begin{lemma}[LS in the Four Subspaces of Linear Algebra via SVD]\label{lemma:fourspace-svd-ls}
Assume $\bX \in \real^{n\times p}$ is fixed and does \textbf{not} necessarily have full rank with $n>p$. Consider the overdetermined system $\by = \bX\bbeta$.
Then, we can split $\by$ into $\hat{\by} + \be$, where $\hat{\by}$ is in the column space of $\bX$ and $\be$ is in the null space of $\bX^\top$. We can always find this split since the column space of $\bX$ and the null space of $\bX^\top$ span the whole space $\real^n$. The relationship between $\be$ and $\bp$ is shown in Figure~\ref{fig:lafundamental4-LS-SVD}. Let $\bX^+ = \bV\bSigma^+\bU^\top$ be the pseudo-inverse of $\bX$. The pseudo-inverse $\bX^+$ agrees with $\bX^{-1}$ when $\bX$ is invertible. Then, we have the following properties (also shown in Figure~\ref{fig:lafundamental4-LS-SVD}):
\begin{description}
\item 1. For $\be\in \nspace(\bX^\top)$, it follows that $\bX^+\be = \bzero \in \real^p$.
\item 2. Given OLS solution $\hat{\bbeta}$ via SVD, it follows that $\bX^+\hat{\by} = \bX^+\by = \hat{\bbeta}$.
\item 3. OLS solution $\hat{\bbeta}$ is in the row space of $\bX$, i.e., it cannot be split into a combination of two components that are in row space of $\bX$ and null space of $\bX$ respectively. This is the reason why $\hat{\bbeta}$ in Figure~\ref{fig:lafundamental4-LS-SVD} is in the row space of $\bX$ rather than in $\real^p$ in general.
\end{description}
\end{lemma}
\end{svgraybox}
\begin{proof}[of Lemma~\ref{lemma:fourspace-svd-ls}]
Since $\be$ is in $\nspace(\bX^\top)$ and perpendicular to $\cspace(\bX)$, and we have shown in Lemma~\ref{lemma:svd-four-orthonormal-Basis}, $\{\bu_1,\bu_2, \cdots,\bu_r\}$ is an orthonormal basis of $\cspace(\bX)$, then the first $r$ components of $\bU^\top\be$ are all zeros. Therefore, $\bX^+\be = \bV\bSigma^+\bU^\top\be=\bzero$ (see Figure~\ref{fig:lafundamental4-LS-SVD} where we transform $\be$ from $\nspace(\bX^\top)$ into the $\bzero \in \real^p$ by $\bX^+$). Moreover, $\hat{\bbeta}=\bX^+\by = \bX^+(\hat{\by}+\be) = \bX^+\hat{\by}$.

Further, we have also shown in Lemma~\ref{lemma:svd-four-orthonormal-Basis} that $\{\bv_1, \bv_2, \cdots, \bv_r\} $ is an orthonormal basis of $\cspace(\bX^\top)$, $\hat{\bbeta} = \sum_{i=1}^{r}\frac{\bu_i^\top \by}{\sigma_i}\bv_i$ thus is in the row space of $\bX$.
\end{proof}
Actually, from projection point of view, the $\hat{\by}$ is the closest point to $\by$ in the column space of $\bX$. This point is the projection $\hat{\by}$ of $\by$. Then the error vector $\be=\by-\hat{\by}$ has minimal length \citep{lu2021rigorous}.

Apart from this LS solution from SVD, in practice, a direct solution of the normal equations can lead to numerical difficulties when $\bX^\top\bX$ is close to singular. In particular, when two or more of the columns in $\bX^\top\bX$ are co-linear, the resulting parameter values can have large magnitude. Such near degeneracies will not be uncommon when dealing with real data sets. The resulting numerical difficulties can be addressed using the SVD as well \citep{bishop2006pattern}.

\section{Pseudo-Inverse in the Fundamental Theorem of Linear Algebra}\label{appendix:pseudo-inverse}
If the matrix $\bX$ is nonsingular, then the linear system $\by = \bX\bbeta$ can be easily solved by the inverse of $\bX$ such that $\hat{\bbeta} = \bX^{-1}\by$. However, the inverse of an $n\times p$ matrix $\bX$ does not exist if $\bX$ is not square or $\bX$ is singular. But we can still find its pseudo-inverse, a $p\times n$ matrix denoted by $\bX^+$.
Before the discussion of pseudo-inverse, we firstly introduce one-sided inverse, generalized inverse, and reflexive generalized inverse that are prerequisites of pseudo-inverse and some the properties of them.
\subsection{One-sided Inverse}
\begin{svgraybox}
	\begin{definition}[One-sided Inverse]
		For any matrix $\bX\in \real^{n\times p}$, if there is a matrix $\bX_L^{-1}$ such that
		$$
		\bX_L^{-1} \bX = \bI_p,
		$$
		then $\bX_L^{-1}$ is a \textbf{left inverse} of $\bX$, and $\bX$ is called \textbf{left-invertible}.
		Similarly, if there is a matrix $\bX_R^{-1}$ such that 
		$$
		\bX \bX_R^{-1}= \bI_n,
		$$
		then $\bX_R^{-1}$ is a \textbf{right inverse} of $\bX$, and $\bX$ is called \textbf{right-invertible}.
		
		\textbf{A word on the notation}: Note here the superscript $-1$ in $\bX_L^{-1}$ and $\bX_R^{-1}$ does not mean the inverse of $\bX_L$ or $\bX_R$ but the one-sided inverse of $\bX$.
	\end{definition}
\end{svgraybox}

\begin{svgraybox}
	\begin{lemma}[One-sided Invertible]\label{theorem:one-sided-invertible}
		For any matrix $\bX\in \real^{n\times p}$, we have 
		\item $\bullet$ $\bX$ is left-invertible if and only if $\bX$ has full column rank (which implies $n\geq p$);
		\item $\bullet$ $\bX$ is right-invertible if and only if $\bX$ has full row rank (which implies $n\leq p$).
	\end{lemma}
\end{svgraybox}
\begin{proof}[of Lemma~\ref{theorem:one-sided-invertible}]
	Suppose $\bX$ has full column rank, then $\bX^\top \bX \in \real^{p\times p}$ has full rank (see Lemma~\ref{lemma:rank-of-ata-x}). Therefore, $(\bX^\top \bX)^{-1}(\bX^\top \bX) = \bI_p$. That is \textcolor{blue}{$(\bX^\top \bX)^{-1}\bX^\top$ is a left inverse of $\bX$}. 
	
	For the reverse, now suppose $\bX$ is left-invertible such that $\bX_L^{-1} \bX = \bI_p$. Since all rows of $\bX_L^{-1} \bX$ are the combinations of the rows of $\bX$, that is, the row space of $\bX_L^{-1} \bX$ is a subset of the row space of $\bX$. We then have $rank(\bX) \geq rank(\bX_L^{-1} \bX) =rank(\bI_p) = p$ which implies $rank(\bX)=p$ and $\bX$ has full column rank. 
	
	Similarly, we can show $\bX$ is right-invertible if and only if $\bX$ has full row rank and \textcolor{blue}{$\bX^\top (\bX\bX^\top)^{-1}$ is a right inverse of $\bX$}.
\end{proof}

We have shown in the above proof that $(\bX^\top \bX)^{-1}\bX^\top$ is a specific left inverse of $\bX$ if $\bX$ has full column rank, and $\bX^\top (\bX\bX^\top)^{-1}$ is a specific right inverse of $\bX$ if $\bX$ has full row rank.
However, the inverse of a $p\times p$ nonsingular matrix requires $2p^3$ floating points operations (flops) which is a complex procedure \citep{lu2021numerical}.
In our case, the inverse of $\bX^\top\bX$ requires $2p^3$ flops and the inverse of $\bX\bX^\top$ requires $2n^3$ flops. A simpler way to find a one-sided inverse is through elementary operations.

Suppose $\bX\in \real^{n\times p}$ has full column rank and we apply \textbf{row elementary operations} $\bE\in \real^{n\times n}$ on $[\bX, \bI_n]$ such that
\begin{equation}\label{equation:onesided-1}
	\bE \begin{bmatrix}
		\bX & \bI_n
	\end{bmatrix}=
	\begin{bmatrix}
		\bI_p & \bG \\
		\bzero & \bZ
	\end{bmatrix},
\end{equation}
where $\bG \in \real^{p\times n}$, $\bI_n$ is an $n\times n$ identity matrix, $\bI_p$ is a $p\times p$ identity matrix, and $\bZ$ is an $(n-p)\times n$ matrix.
Then, it can be easily verified that $\bG\bX = \bI_p$ and $\bG$ is a left inverse of $\bX$.

Similarly, suppose $\bX\in \real^{n\times p}$ has full row rank and we apply \textbf{column elementary operations} $\bE \in \real^{p\times p}$ on $[\bX^\top, \bI_p]^\top$ such that 
\begin{equation}\label{equation:onesided-2}
	\begin{bmatrix}
		\bX \\
		\bI_p
	\end{bmatrix} \bE=
	\begin{bmatrix}
		\bI_n & \bzero\\
		\bG & \bZ
	\end{bmatrix},
\end{equation}
where $\bZ$ is a $p\times (p-n)$ matrix.
Then, $\bG \in \real^{p\times n}$ is a right inverse of $\bX$.

More generally, the following two propositions show us how to find more left inverses or right inverses of a matrix.
\begin{svgraybox}
	\begin{proposition}[Finding Left Inverse]
		Suppose $\bX\in \real^{n\times p}$ is left-invertible ($n\geq p$), then 
		$$
		\bX_L^{-1} = \begin{bmatrix}
			(\bX_1^{-1} -\bY \bX_2 \bX_1^{-1}) & \bY
		\end{bmatrix}\bE,
		$$
		is a left inverse of $\bX$, where $\bY\in \real^{p\times (n-p)}$ can be any matrix, and $\bE\bX = \begin{bmatrix}
			\bX_1 \\
			\bX_2
		\end{bmatrix}$ is the row elementary transformation of $\bX$ such that $\bX_1 \in \real^{p\times p}$ is invertible (since $\bX$ has full column rank $p$) and $\bE\in \real^{n\times n}$.
	\end{proposition}
\end{svgraybox}
We can verify that $\bG$ in Equation~\ref{equation:onesided-1} is a specific left inverse of $\bX$ by setting $\bY=\bzero$. Since $\bE = \begin{bmatrix}
	\bG \\
	*
\end{bmatrix}$, $\bX_1 = \bI_p$, and $\bX_2 = \bzero$, we have 
$$
\bX_L^{-1} = \begin{bmatrix}
	(\bX_1^{-1} -\bY \bX_2 \bX_1^{-1}) & \bY
\end{bmatrix}\bE = \bG + \bY\bZ = \bG,
$$
where the last equation is from the assumption that $\bY=\bzero$.

\begin{svgraybox}
	\begin{proposition}[Finding Right Inverse]
		Suppose $\bX\in \real^{n\times p}$ is right-invertible ($n\leq p$), then 
		$$
		\bX_R^{-1} = \bE\begin{bmatrix}
			(\bX_1^{-1}-\bX_1^{-1}\bX_2\bY) \\
			\bY
		\end{bmatrix},
		$$
		is a right inverse of $\bX$, where $\bY\in \real^{(p-n)\times n}$ can be any matrix, and $\bX\bE = \begin{bmatrix}
			\bX_1 &\bX_2 
		\end{bmatrix}$ is the column elementary transformation of $\bX$ such that $\bX_1 \in \real^{n\times n}$ is invertible (since $\bX$ has full row rank $n$) and $\bE\in \real^{p\times p}$.
	\end{proposition}
\end{svgraybox}
Similarly, we can verify that $\bG$ in Equation~\ref{equation:onesided-2} is a specific right inverse of $\bX$ by setting $\bY=\bzero$. Since $\bE = [\bG, \bZ]$, $\bX_1=\bI_n$, and $\bX_2=\bzero$, we have 
$$
\bX_R^{-1} = \bE\begin{bmatrix}
	(\bX_1^{-1}-\bX_1^{-1}\bX_2\bY) \\
	\bY
\end{bmatrix} = \bG+\bZ\bY = \bG,
$$
where again the last equality is from the assumption that $\bY=\bzero$.

\begin{figure}[h!]
	\centering
	\includegraphics[width=0.98\textwidth]{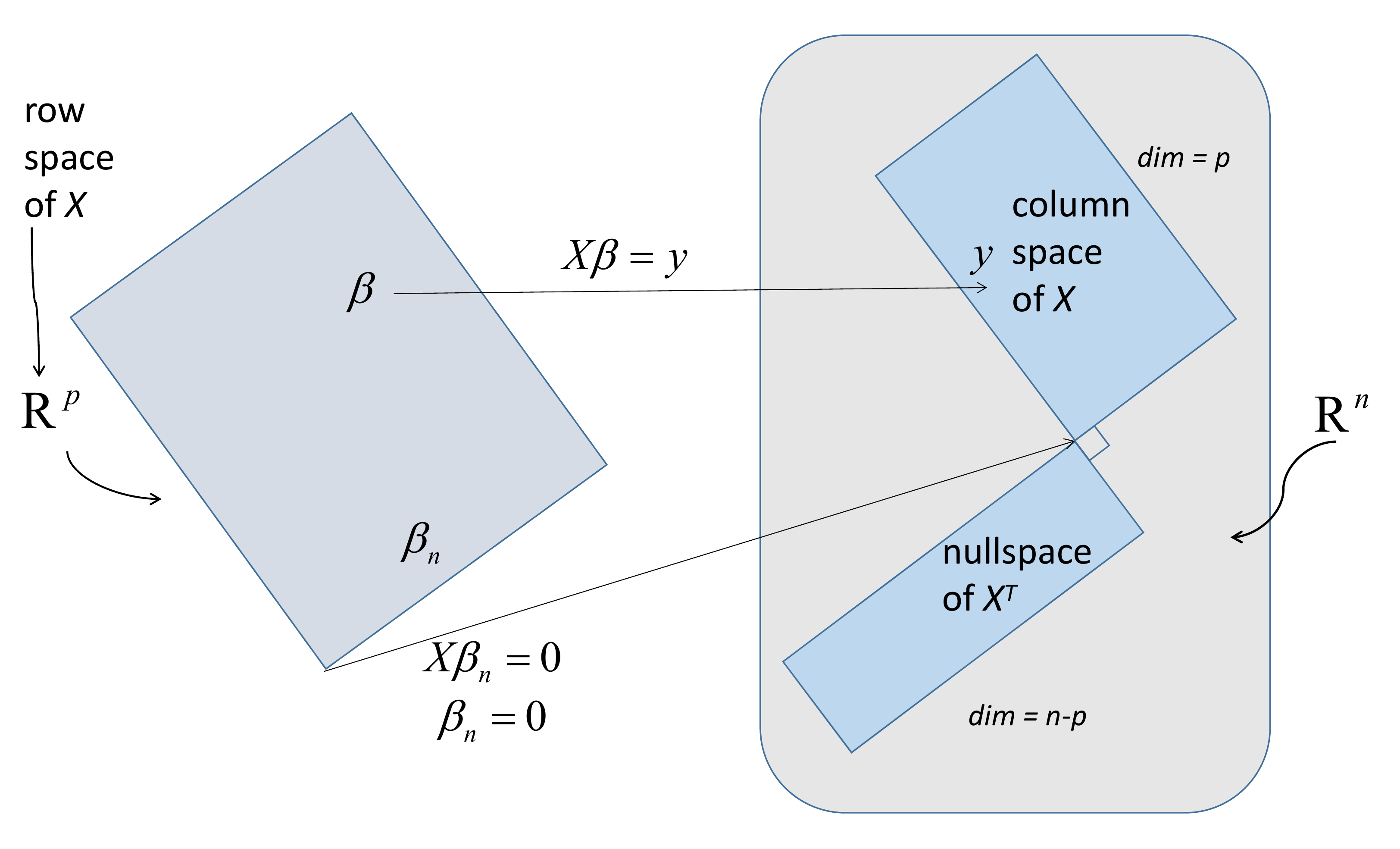}
	\caption{\textbf{Fifth Figure}: $\bX$ is left-invertible, the row space of $\bX$ is the whole space of $\real^p$. $(\bI_n - \bX \bX_L^{-1})\by = \bzero$ implies $\by$ is in the column space of $\bX$ such that the solution of $\bX\bbeta=\by$ is unique.}
	\label{fig:lafundamental-left-inverse}
\end{figure}
Under specific condition, the linear system $\bX\bbeta=\by$ has an unique solution.
\begin{svgraybox}
	\begin{proposition}[Unique Linear System Solution]\label{proposition:unique-linear-system-solution}
		Suppose $\bX\in \real^{n\times p}$ is left-invertible ($n\geq p$), and $\bX_L^{-1}$ is the left inverse of $\bX$. Then the linear system $\bX\bbeta = \by$ has an \textbf{unique} solution if and only if 
		$$
		(\bI_n - \bX \bX_L^{-1})\by = \bzero.
		$$
		And the unique solution is given by 
		$$
		\hat{\bbeta} = (\bX^\top\bX)^{-1}\bX^\top \by.
		$$
	\end{proposition}
\end{svgraybox}
\begin{proof}[of Proposition~\ref{proposition:unique-linear-system-solution}]
	Suppose $\bbeta_0$ is the solution of $\bX\bbeta =\by$, then 
	$$
	\begin{aligned}
		\bX \bX_L^{-1} (\bX \bbeta_0) &= \bX\bX_L^{-1} \by \\
		\bX (\bX_L^{-1} \bX) \bbeta_0 &= \bX \bbeta_0 = \by.
	\end{aligned}
	$$
	That is, $\bX\bX_L^{-1} \by=\by$ and $(\bI_n-\bX\bX_L^{-1})\by = \bzero$. 
	For the reverse, suppose $(\bI_n - \bX \bX_L^{-1})\by = \bzero$, and let $\bbeta_0 = \bX_L^{-1} \by$. Then substitute $\bbeta_0 = \bX_L^{-1} \by$ into $(\bI_n - \bX \bX_L^{-1})\by = \bzero$, we have 
	$$
	\bX\bbeta_0 = \by,
	$$
	which implies $\bbeta_0 = \bX_L^{-1} \by$ is a solution of $\bX\bbeta = \by$ if $(\bI_n - \bX \bX_L^{-1})\by = \bzero$.
	
	To prove the uniqueness, suppose $\bbeta_0$ and $\bbeta_1$ are two solutions of $\bX\bbeta = \by$. We have $\bX\bbeta_0=\bX\bbeta_1 = \by$ such that $\bX(\bbeta_0-\bbeta_1) = \bzero$. Since $\bX$ is left-invertible so that $\bX$ has full column rank $p$, the dimension of the row space of $\bX$ is $p$ as well such that the null space of $\bX$ is of dimension 0 (i.e., $\dim(\cspace(\bX^\top)) + \dim(\nspace(\bX))=p$ by fundamental theorem of linear algebra, see Theorem~\ref{theorem:fundamental-linear-algebra}). Then $\bbeta_0=\bbeta_1$ which completes the proof.
\end{proof}

In the fundamental theorem of linear algebra Figure~\ref{fig:lafundamental-ls}, \textcolor{blue}{the row space of $\bX$ is the whole space of $\real^p$ if $\bX$ is left-invertible} (i.e., $\bX$ has full column rank $p$). If the condition $(\bI_n - \bX \bX_L^{-1})\by = \bzero$ is satisfied, it implies that $\by$ is in the column space of $\bX$ such that $\bX\bbeta=\by$ has at least one solution and the above proposition shows that this solution is unique. The situation is shown in Figure~\ref{fig:lafundamental-left-inverse}.

\begin{figure}[h!]
	\centering
	\includegraphics[width=0.98\textwidth]{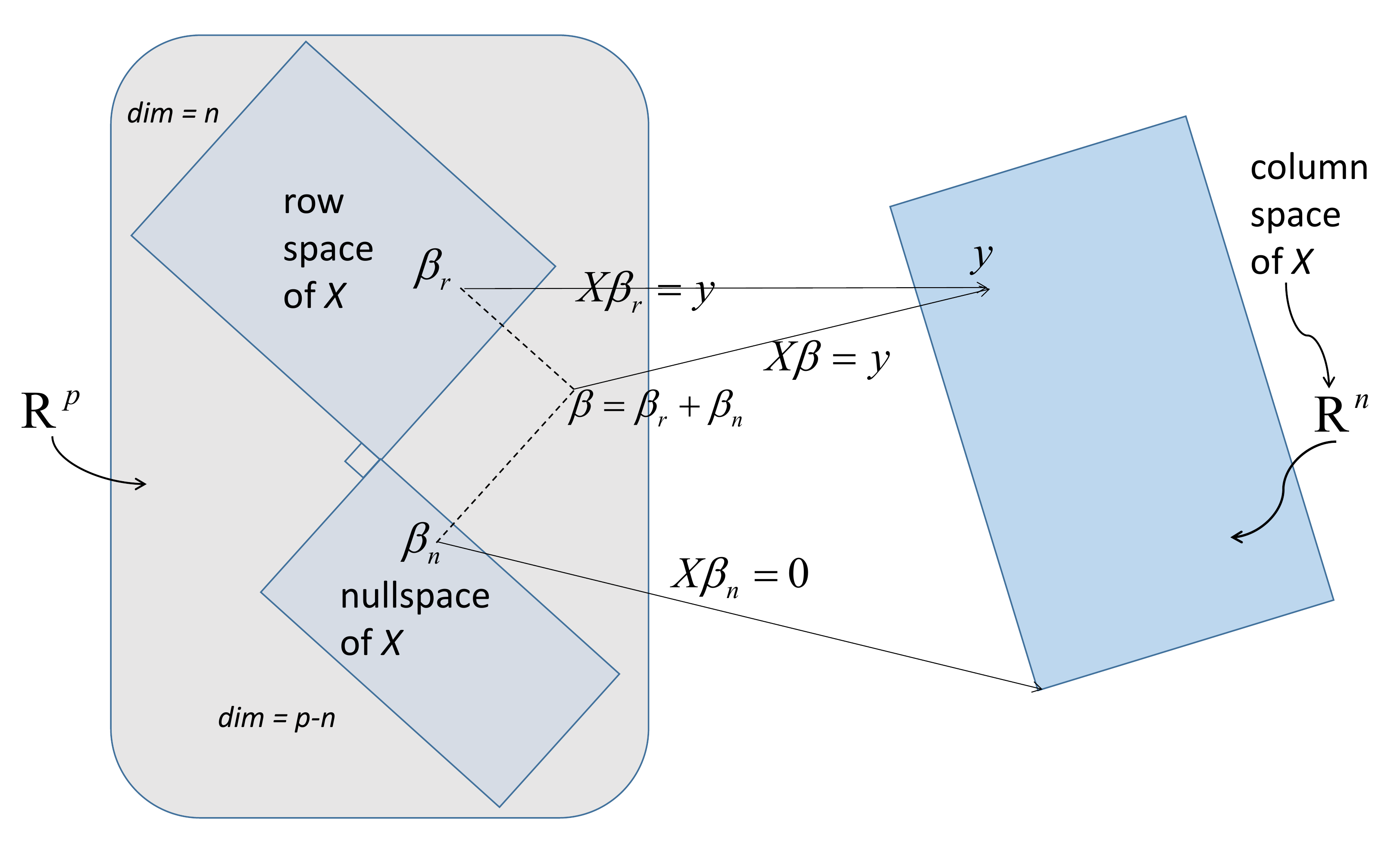}
	\caption{\textbf{Sixth Figure}: $\bX$ is right-invertible, the column space of $\bX$ is the whole space of $\real^n$.}
	\label{fig:lafundamental-right-inverse}
\end{figure}
\begin{svgraybox}
	\begin{proposition}[Always Have Solution]\label{proposition:always-have-solution-right-inverse}
		Suppose $\bX\in \real^{n\times p}$ is right-invertible (which implies $n\leq p$), and $\bX_R^{-1}$ is a right inverse of $\bX$. Then for any $\by\in \real^n$, the linear system $\bX\bbeta = \by$ has solutions, and the solution is given by
		$$
		\hat{\bbeta} = \bX_R^{-1}\by, 
		$$
		where $\bX_R^{-1}$ is a right inverse of $\bX$ and the right inverse is not necessarily unique.
	\end{proposition}
\end{svgraybox}
\begin{proof}[of Proposition~\ref{proposition:always-have-solution-right-inverse}]
	It can be easily verified that 
	$$
	(\bX \bX_R^{-1}) \by = \bI_n \by =\by,
	$$
	so that $\bX_R^{-1}\by$ is a solution of $\bX\bbeta=\by$.
\end{proof}

We notice that if $\bX$ is right-invertible, then $\bX$ has full row rank $n$. In the fundamental theorem of linear algebra Figure~\ref{fig:lafundamental-ls}, \textcolor{blue}{the column space of $\bX$ is the whole space of $\real^n$ if $\bX$ is right-invertible}. Then any vector $\by\in \real^n$ is in the column space of $\bX$ such that $\bX\bbeta=\by$ has at least one solution. The situation is shown in Figure~\ref{fig:lafundamental-right-inverse}. This implies $p\geq n$. When $p >n$, we can always find some $\bbeta_n\in \nspace(\bX)$ such that the solution of $\bX\bbeta=\by$ is not unique. When $p=n$, the solution is apparently unique and the right inverse degenerates to the inverse.

\subsection{Generalized Inverse (g-inverse)}
We mentioned previously that if matrix $\bX$ is nonsingular, then the linear system $\by = \bX\bbeta$ can be easily solved by the inverse of $\bX$ such that $\hat{\bbeta} = \bX^{-1}\by$. However, the inverse of an $n\times p$ matrix $\bX$ does not exist if $\bX$ is not square or $\bX$ is singular. But still, when $\by$ is in the column space of $\bX$, we can find the solution of the linear system as well. The relationship between the solution $\hat{\bbeta}$ and $\by$ is given by the generalized inverse of $\bX$: $\hat{\bbeta} = \bX^-\by$.
\begin{svgraybox}
	\begin{definition}[Generalized Inverse]
		Any matrix $\bX \in \real^{n\times p}$ has rank $r$ with $r \leq p \leq n$. Then a generalized inverse $\bX^- \in\real^{p\times n}$ of $\bX$ is a matrix that satisfies
		$$
		(C1) \qquad  \bX\bX^-\bX = \bX,
		$$
		or equivalently, 
		$$
		(C1.1) \qquad \bX\bX^-\by = \by
		$$
		for any $\by \in \cspace(\bX)$.
	\end{definition}
\end{svgraybox}
To see the equivalence between $(C1)$ and $(C1.1)$, that is, we want to show $\bX$ satisfies $(C1)$ if and only if it satisfies $(C1.1)$. For any $\by \in \cspace(\bX)$, we can find a $\bbeta\in\real^p$ such that $\bX\bbeta = \by$. If $\bX$ and $\bX^-$ satisfy $(C1)$, then
$$
\bX\bX^-\bX \bbeta = \bX\bbeta \leadto \bX\bX^-\by = \by,
$$
which implies $\bX$ and $\bX^-$ also satisfy $(C1.1)$. For the reverse, suppose $\bX$ and $\bX^-$ satisfy $(C1.1)$, then 
$$
\bX\bX^-\by = \by \leadto \bX\bX^-\bX \bbeta = \bX\bbeta ,
$$
which implies $\bX$ and $\bX^-$ also satisfy $(C1)$.

Multiply on the left of $(C1)$ by $\bX^-$ and follow the definition of projection matrix,\footnote{A \textbf{projection matrix} is a matrix that is idempotent. And an \textbf{orthogonal projection matrix} is a matrix that is both symmetric and idempotent. See \citep{lu2021rigorous}.} we obtain $\bX^-\bX\bX^-\bX = \bX^-\bX$ such that $ \bX^-\bX$ is idempotent, which implies $\bX^-\bX$ is a projection matrix (not necessarily to be an orthogonal projection).
\begin{svgraybox}
	\begin{lemma}[Projection Matrix from Generalized Inverse]\label{lemma:idempotent-of-ginverse}
		For any matrix $\bX$, and its generalized inverse $\bX^-$, $\bX^-\bX$ is a projection matrix but not necessarily an orthogonal projection. Same claim can be applied to $\bX\bX^-$ as well.
	\end{lemma}
\end{svgraybox}

\begin{svgraybox}
	\begin{lemma}[Rank of Generalized Inverse]\label{proposition:rank-of-ginverse}
		For any matrix $\bX \in \real^{n\times p}$, and its generalized inverse $\bX^- \in\real^{p\times n}$, then 
		$$
		rank(\bX^-) \geq rank(\bX).
		$$
		Specifically, we also have $rank(\bX)=rank(\bX\bX^-)=rank(\bX^-\bX)$.
	\end{lemma}
\end{svgraybox}
\begin{proof}[of Lemma~\ref{proposition:rank-of-ginverse}]
	From $(C1)$, we notice that $rank(\bX) = rank( \bX\bX^-\bX)$. And
	$$
	rank( \bX\bX^-\bX) \leq rank(\bX\bX^-)\leq rank(\bX^-),
	$$
	where the first inequality comes from the fact that the columns of $\bX\bX^-\bX$ are combinations of columns of $\bX\bX^-$, and the second inequality comes from the fact that the rows of $\bX\bX^-$ are combinations of rows of $\bX^-$.
	
	For the second part, we have 
	$$
	rank(\bX) \geq rank(\bX\bX^-) \geq rank(\bX\bX^-\bX),
	$$
	where the first inequality is from the fact that the columns of $\bX\bX^-$are combinations of the columns of $\bX$, and the second inequality is from the fact that the columns of $\bX\bX^-\bX$ are combinations of the columns of $\bX\bX^-$. From $(C1)$ again, $rank(\bX) = rank( \bX\bX^-\bX)$ which implies by "sandwiching" that
	$$
	rank(\bX) = rank(\bX\bX^-) = rank(\bX\bX^-\bX).
	$$ 
	Similarly, we also have 
	$$
	rank(\bX) \geq rank(\bX^-\bX) \geq rank(\bX\bX^-\bX),
	$$
	where the first inequality is from the fact that the rows of $\bX^-\bX$ are combinations of the rows of $\bX$, and the second inequality is from the fact that the rows of $\bX\bX^-\bX$ are combinations of the rows of $\bX^-\bX$. By "sandwiching" again, we have 
	$$
	rank(\bX) = rank(\bX^-\bX) = rank(\bX\bX^-\bX),
	$$ 
	which completes the proof.
\end{proof}

In Lemma~\ref{theorem:one-sided-invertible}, we have shown that the left inverse exists if and only if $\bX$ has full column rank, and the right inverse exists if and only if $\bX$ has full row rank. However, this is not required in generalized inverse. When this full rank condition is satisfied, we have another property as follows.
\begin{svgraybox}
	\begin{lemma}[Full Rank Generalized Inverse]\label{Lemma:fullrank-ginverse}
		For any matrix $\bX \in \real^{n\times p}$, and its generalized inverse $\bX^- \in\real^{p\times n}$, then we have 
		\item 1). $\bX$ has full column rank if and only if $\bX^-\bX = \bI_p$;
		\item 2). $\bX$ has full row rank if and only if $\bX\bX^- = \bI_n$.
	\end{lemma}
\end{svgraybox}
\begin{proof}[of Lemma~\ref{Lemma:fullrank-ginverse}]
	For 1). Suppose $\bX$ has full column rank, and we have shown in Lemma~\ref{proposition:rank-of-ginverse} that $rank(\bX)=rank(\bX\bX^-)=rank(\bX^-\bX)$. Then $rank(\bX^-\bX)=rank(\bX)=p$ and $\bX^-\bX\in \real^{p\times p}$ is nonsingular. We have 
	$$
	\bI_p= (\bX^-\bX)(\bX^-\bX)^{-1} = \bX^-(\bX\bX^-\bX)(\bX^-\bX)^{-1} = \bX^-\bX.
	$$
	For the reverse, suppose $\bX^-\bX = \bI_p$ which implies $rank(\bX^-\bX)=p$. From $rank(\bX^-\bX)=rank(\bX)$, we have $rank(\bX)=p$ such that $\bX$ has full column rank.
	
	Similarly, we can show $\bX$ has full row rank if and only if $\bX\bX^- = \bI_n$.
\end{proof}

\begin{svgraybox}
	\begin{lemma}[Constructing Generalized Inverse]\label{Lemma:construct-ginverse}
		For any matrix $\bX \in \real^{n\times p}$, and its generalized inverse $\bX^- \in\real^{p\times n}$, then there exists a $p\times n$ matrix $\bA$ such that 
		\begin{equation}\label{equation:constructing-ginverse}
			\bar{\bX} = \bX^-+ \bA-\bX^-\bX\bA\bX\bX^-
		\end{equation}
		is also a generalized inverse of $\bX$. In addition, for any generalized inverse $\bar{\bX}$, there exists a matrix $\bA$ so that Equation~\ref{equation:constructing-ginverse} is satisfied.
	\end{lemma}
\end{svgraybox}
\begin{proof}[of Lemma~\ref{Lemma:construct-ginverse}]
	Write out the equation
	$$
	\begin{aligned}
		\bX\bar{\bX}\bX &= \bX(\bX^-+ \bA-\bX^-\bX\bA\bX\bX^-)\bX = \bX\bX^-\bX + \bX\bA\bX  -(\bX\bX^-\bX)\bA(\bX\bX^-\bX)\\
		&= \bX\bX^-\bX + \bX\bA\bX  -\bX\bA\bX = \bX,
	\end{aligned}
	$$
	so that $\bar{\bX}$ is a generalized inverse of $\bX$. 
	
	Suppose now that $\bB$ is any generalized inverse of $\bX$, and define $\bA = \bB-\bX^-$. Recall that $\bX\bB\bX = \bX$, we have 
	$$
	\begin{aligned}
		\bX^-+ \bA-\bX^-\bX\bA\bX\bX^- &= \bX^-+ (\bB-\bX^-)-\bX^-\bX(\bB-\bX^-)\bX\bX^- \\
		&= \bB - \bX^-(\bX\bB\bX)\bX^- + \bX^-(\bX\bX^-\bX)\bX^-\\
		&= \bB - \bX^-\bX\bX^- + \bX^-\bX\bX^- \\
		&= \bB,
	\end{aligned}
	$$
	which implies $\bA$ can be constructed for any generalized inverse $\bB$.
\end{proof}

\begin{svgraybox}
	\begin{lemma}[Generalized Inverse Properties]\label{proposition:ginverse-properties}
		For any matrix $\bX \in \real^{n\times p}$, and its generalized inverse $\bX^- \in\real^{p\times n}$, then 
		\item 1). $(\bX^\top)^- = (\bX^-)^\top$, i.e., $(\bX^-)^\top$ is the generalized inverse of $\bX^\top $;
		\item 2). For any $a\neq0$, $\frac{1}{a} \bX^-$ is the generalized inverse of $a\bX$;
		\item 3). Suppose $\bA\in \real^{n\times n}$ and $\bB\in \real^{p\times p}$ are both invertible, then $\bB^{-1}\bX^- \bA^{-1}$ is a generalized inverse of $\bA\bX\bB$;
		\item 4). $\cspace(\bX\bX^-) = \cspace(\bX)$ and $\nspace(\bX^-\bX) = \nspace(\bX)$.
	\end{lemma}
\end{svgraybox}
\begin{proof}[of Lemma~\ref{proposition:ginverse-properties}]
	For 1), from $(C1)$, $\bX\bX^-\bX = \bX$, we have $\bX^\top (\bX^-)^\top\bX^\top =\bX^\top $ such that $(\bX^-)^\top$ is the generalized inverse of $\bX^\top $.
	
	For 2), it can be easily verified that $(a\bX) (\frac{1}{a} \bX^-) (a\bX) = (a\bX)$ such that $\frac{1}{a} \bX^-$ is the generalized inverse of $a\bX$ for any $a\neq 0$.
	
	For 3), we realize that $(\bA\bX\bB) (\bB^{-1}\bX^- \bA^{-1})(\bA\bX\bB)= \bA\bX\bX^- \bX\bB=\bA\bX\bB$ which implies $\bB^{-1}\bX^- \bA^{-1}$ is a generalized inverse of $\bA\bX\bB$.
	
	For 4), since the columns of $\bX\bX^-$ are the combinations of the columns of $\bX$, then $\cspace(\bX\bX^-) \subseteq \cspace(\bX)$. And we proved that $rank(\bX)=rank(\bX\bX^-)$ in Lemma~\ref{proposition:rank-of-ginverse}, then $\cspace(\bX\bX^-) = \cspace(\bX)$. Similarly, we could prove $\nspace(\bX^-\bX) = \nspace(\bX)$.

\end{proof}

\subsection{Reflexive Generalized Inverse (rg-inverse)}

\begin{figure}[h!]
	\centering
	\includegraphics[width=0.8\textwidth]{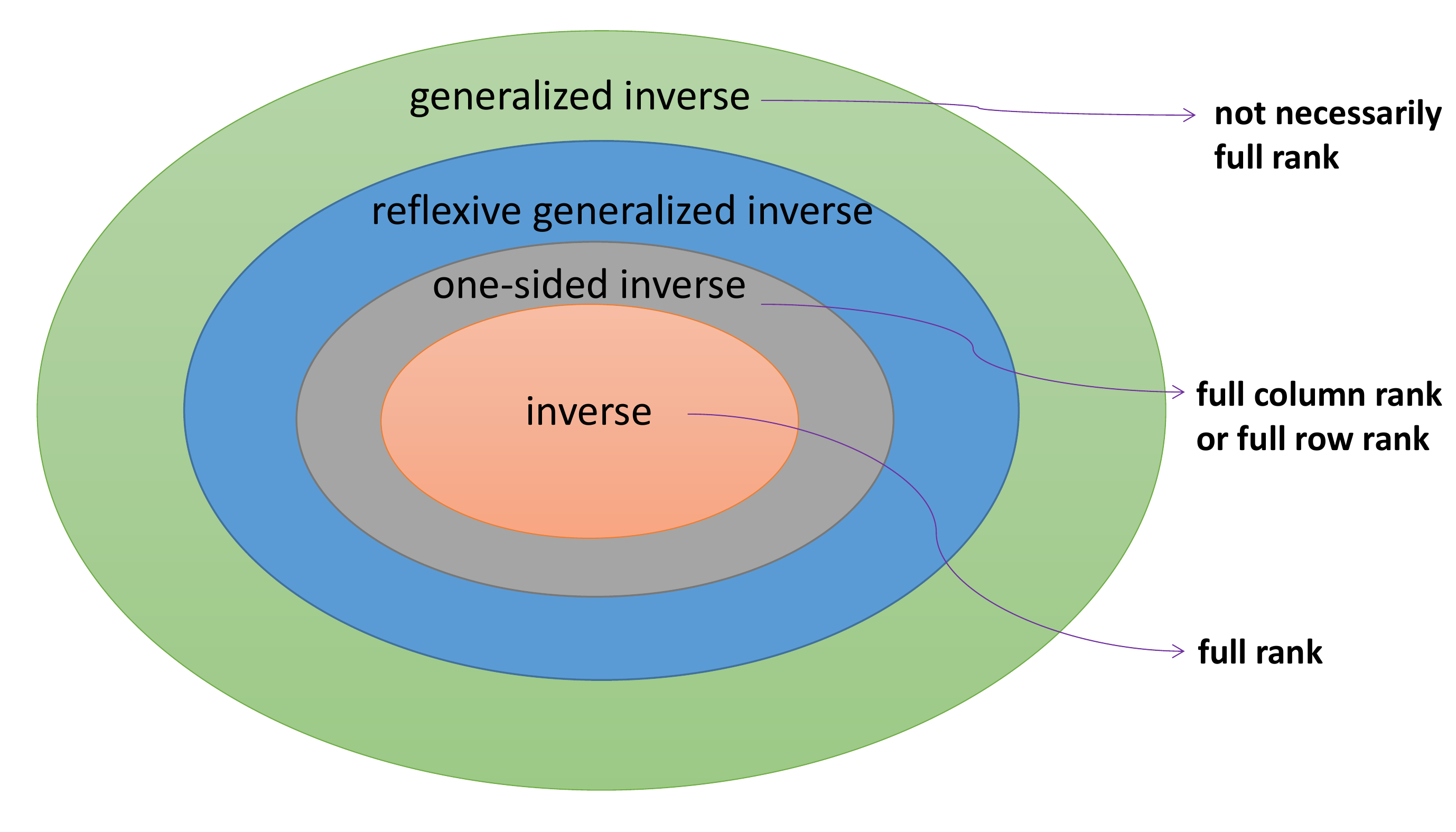}
	\caption{Relationship of different inverses.}
	\label{fig:pseudo-inverse-comparison}
\end{figure}
\begin{svgraybox}
	\begin{definition}[Reflexive Generalized Inverse]
		Any matrix $\bX \in \real^{n\times p}$ has rank $r$ with $r \leq \min\{p, n\}$. Then a reflexive generalized inverse $\bX_r^- \in\real^{p\times n}$ of $\bX$ is a matrix that satisfies
		$$
		(C1) \qquad  \bX\bX_r^-\bX = \bX,
		$$
		and
		$$
		(C2) \qquad \bX_r^- \bX \bX_r^- = \bX_r^-.
		$$ 
		That is $\bX_r^-$ is a g-inverse of $\bX$, and $\bX$ is a g-inverse of $\bX_r^-$.
	\end{definition}
\end{svgraybox}
Let $\bX$ be of rank $r$, it can be factored as $\bX = \bE_1 \begin{bmatrix}
	\bI_r & \bzero\\
	\bzero & \bzero
\end{bmatrix}\bE_2$, where $\bE_1\in \real^{n\times n}$, $\bE_2\in \real^{p\times p}$ are elementary transformations on $\bX$. Then, we can construct such a reflexive generalized inverse of $\bX$ as 
$$
\bX_r^- = \bE_2^{-1} \begin{bmatrix}
	\bI_r & \bA \\
	\bB  & \bB\bA
\end{bmatrix}\bE_1^{-1},
$$
where $\bA\in \real^{r\times (n-r)}$, $\bB\in \real^{(p-r)\times r}$ can be any matrix so that the reflexive generalized inverse is \textbf{not unique}. This construction of the reflexive generalized inverse also shows that  reflexive generalized inverse exists for any matrix. This implies reflexive generalized inverse is a more general inverse of $\bX$ compared to the one-sided inverse which may not exist.
\begin{svgraybox}
	\begin{lemma}[Reflexive Generalized Inverse from G-Inverse]\label{lemma:generalized-for-reflexive-two}
		For any matrix $\bX \in \real^{n\times p}$, and $\bA$, $\bB$ are two generalized inverses of $\bX$, then 
		$$
		\bZ = \bA\bX\bB 
		$$
		is a reflexive generalized inverse of $\bX$.
	\end{lemma}
\end{svgraybox}
It can be easily verified that $\bX\bZ\bX = \bX$ and $\bZ\bX\bZ = \bZ$.

\begin{svgraybox}
	\begin{lemma}[Reflexive Generalized Inverse from G-Inverse]\label{lemma:generalized-for-reflexive-two22}
		For any matrix $\bX \in \real^{n\times p}$, then the following two matrices $\bA$ and $\bB$ are two reflexive generalized inverses of $\bX$:
		$$
		\begin{aligned}
			\bA &=(\bX^\top\bX)^-\bX^\top, \\
			\bB &=\bX^\top (\bX\bX^\top)^-,
		\end{aligned}
		$$
		where $(\bX^\top\bX)^-$ is a g-inverse of $(\bX^\top\bX)$, and $(\bX\bX^\top)^-$ is a g-inverse of $(\bX\bX^\top)$.
	\end{lemma}
\end{svgraybox}
\begin{proof}[of Lemma~\ref{lemma:generalized-for-reflexive-two22}] 
	Let's first check the following result.
	\begin{mdframed}[hidealllines=\mdframehidelineNote,backgroundcolor=\mdframecolorSkip,frametitle={$\cspace(\bX^\top \bX) = \cspace(\bX^\top) \qquad \text{and} \qquad \nspace(\bX^\top \bX)=\nspace(\bX)$}]
		Since the columns of $\bX^\top \bX$ are combinations of the columns of $\bX^\top$, we have $\cspace(\bX^\top \bX) \subseteq \cspace(\bX^\top)$.
		In Lemma~\ref{lemma:rank-of-ata-x}, we proved that, $rank(\bX^\top \bX)=rank(\bX)$. This implies $rank(\bX^\top \bX)=rank(\bX^\top)$ and $\cspace(\bX^\top \bX) = \cspace(\bX^\top)$. Furthermore, the orthogonal complement of $\cspace(\bX^\top)$ is $\nspace(\bX)$, and the orthogonal complement of $\cspace(\bX^\top \bX)$ is $\nspace(\bX^\top \bX)$. Therefore, by fundamental theorem of linear algebra in Appendix~\ref{appendix:fundamental-rank-nullity}, we have
		$$
		\cspace(\bX^\top \bX) = \cspace(\bX^\top), \qquad \nspace(\bX^\top \bX)=\nspace(\bX).
		$$
	\end{mdframed}
	Then there exists a set of vectors $\bz_1, \bz_2, \cdots, \bz_n \in \real^p$ such that column-$i$ of $\bX^\top$ can be expressed as $\bX^\top\bX\bz_i$. That is, for $\bZ=[\bz_1, \bz_2, \cdots, \bz_n]$, we have
	$$
	\bX^\top = \bX^\top\bX\bZ.
	$$
	Then, 
	$$
	\bX\bA\bX = (\bX^\top\bX\bZ)^\top  (\bX^\top\bX)^-\bX^\top \bX = \bZ^\top \bX^\top\bX (\bX^\top\bX)^-\bX^\top\bX.
	$$
	By condition $(C1.1)$ of g-inverse, we have $\bX^\top\bX (\bX^\top\bX)^- \by = \by $ for any $\by$. This implies  $\bX^\top\bX (\bX^\top\bX)^-\bX^\top\bX = \bX^\top\bX$ and 
	\begin{equation}\label{equation:reflex-eq1}
		\bX\bA\bX = (\bX^\top\bX\bZ)^\top  (\bX^\top\bX)^-\bX^\top \bX = \bZ^\top \bX^\top\bX=\bX.
	\end{equation}
	Write out $\bA\bX\bA$, we have
	$$
	\bA\bX\bA = (\bX^\top\bX)^-\bX^\top \bX(\bX^\top\bX)^-\bX^\top.
	$$
	Same argument can be applied to $\bX^\top \bX(\bX^\top\bX)^-\bX^\top = \bX^\top$. Then,
	\begin{equation}\label{equation:reflex-eq2}
		\bA\bX\bA = (\bX^\top\bX)^-\bX^\top = \bA.
	\end{equation}
	Combine Equation~\ref{equation:reflex-eq1} and Equation~\ref{equation:reflex-eq2}, we conclude that $\bA$ is a reflexive generalized inverse of $\bX$. Similarly, we can show $\bB$ is a reflexive generalized inverse of $\bX$ as well.
\end{proof}

From the definition, we realize that reflexive generalized inverse is a special generalized inverse. Under specific condition, the two inverses are equivalent.

\begin{svgraybox}
	\begin{lemma}[Reflexive Generalized Inverse in G-Inverse]\label{lemma:reflexive-from-ginverse}
		For any matrix $\bX \in \real^{n\times p}$, and $\bX^-\in \real^{p\times n}$ is a generalized inverse of $\bX$, then $\bX^-$ is a reflexive generalized inverse of $\bX$ if and only if $rank(\bX)=rank(\bX^-)$.
		
	\end{lemma}
\end{svgraybox}
\begin{proof}[of Lemma~\ref{lemma:reflexive-from-ginverse}]
	Suppose $\bX^-$ is a generalized inverse of $\bX$, then $\bX\bX^-\bX=\bX$. Suppose further, $\bX^-$ is also a reflexive generalized inverse, then $\bX^-\bX\bX^- = \bX^-$. We have 
	$$
	\begin{aligned}
		rank(\bX) &= rank(\bX\bX^-\bX)\leq rank(\bX^-) = rank(\bX^-\bX\bX^-)\leq rank(\bX)\\
	\end{aligned}
	$$
	where the two inequalities are from Lemma~\ref{proposition:rank-of-ginverse}. This implies $rank(\bX)=rank(\bX^-)$. 
	
	For the reverse, suppose $\bX^-$ is a generalized inverse of $\bX$, then $\bX\bX^-\bX=\bX$. And suppose further $rank(\bX)=rank(\bX^-)$, we have
	$$
	rank(\bX) = rank(\bX\bX^-\bX) \leq rank(\bX^-\bX) \leq rank(\bX^-) = rank(\bX),
	$$
	where the first inequality is from that fact that the rows of $\bX\bX^-\bX$ are combinations of the rows of $\bX^-\bX$, and the second inequality is from the fact that the columns of $\bX^-\bX$ are combinations of the columns of $\bX^-$. 
	This implies $rank(\bX^-\bX) = rank(\bX^-)$ and $\cspace(\bX^-\bX)=\cspace(\bX^-)$. Then there exists a set of vectors $\balpha_1, \balpha_2, \cdots, \balpha_n \in \real^p$ such that column-$i$ of $\bX^-$ can be expressed as $\bX^-\bX\balpha_i$. That is, for $\bA=[\balpha_1, \balpha_2, \cdots, \balpha_n]$, we have
	$$
	\bX^- = \bX^-\bX\bA.
	$$
	We realize again that $\bX=\bX\bX^-\bX$, then
	$$
	\bX=\bX\bX^-\bX = \bX(\bX^-\bX\bA)\bX = \bX\bA\bX,
	$$
	where the last equality is form condition $(C1.1)$ and $\bA$ is a generalized inverse of $\bX$. From Lemma~\ref{lemma:generalized-for-reflexive-two}, $\bX^-=\bX^-\bX\bA$ is a reflexive generalized inverse of $\bX$ which completes the proof.
\end{proof}
\begin{mdframed}[hidealllines=\mdframehidelineNote,backgroundcolor=\mdframecolorNote]
	\begin{proposition}[Rank of Reflexive Generalized Inverse]
		For any matrix $\bX \in \real^{n\times p}$, and its generalized inverse $\bX_r^- \in\real^{p\times n}$.
		Combine the result in Lemma~\ref{lemma:reflexive-from-ginverse} and the result from the rank of g-inverses in Lemma~\ref{proposition:rank-of-ginverse}, we have
		$$
		rank(\bX_r^-)=rank(\bX)=rank(\bX\bX_r^-)=rank(\bX_r^-\bX).
		$$
	\end{proposition}
\end{mdframed}

\begin{svgraybox}
	\begin{lemma}[Reflexive Generalized Inverse Properties]\label{Lemma:refle-ginverse-properties}
		For any matrix $\bX \in \real^{n\times p}$, and its reflexive generalized inverse $\bX_r^- \in\real^{p\times n}$, then
		\item 1. $\cspace(\bX\bX_r^-) = \cspace(\bX)$ and $\nspace(\bX_r^-\bX) = \nspace(\bX)$.
		\item 2. $\cspace(\bX_r^-\bX) = \cspace(\bX_r^-)$ and $\nspace(\bX\bX_r^-) = \nspace(\bX_r^-)$. 
	\end{lemma}
\end{svgraybox}

\begin{proof}[of Lemma~\ref{Lemma:refle-ginverse-properties}]
	Suppose $\bX^-$ is a g-inverse of $\bX$,
	we proved in Lemma~\ref{proposition:ginverse-properties} that $\cspace(\bX\bX^-) = \cspace(\bX)$ and $\nspace(\bX^-\bX) = \nspace(\bX)$. Since $\bX^-_r$ is a g-inverse of $\bX$, and $\bX$ is a g-inverse of $\bX^-_r$, we complete the proof.
\end{proof}

\subsection{Pseudo-Inverse}
\begin{figure}[h!]
	\centering
	\includegraphics[width=0.98\textwidth]{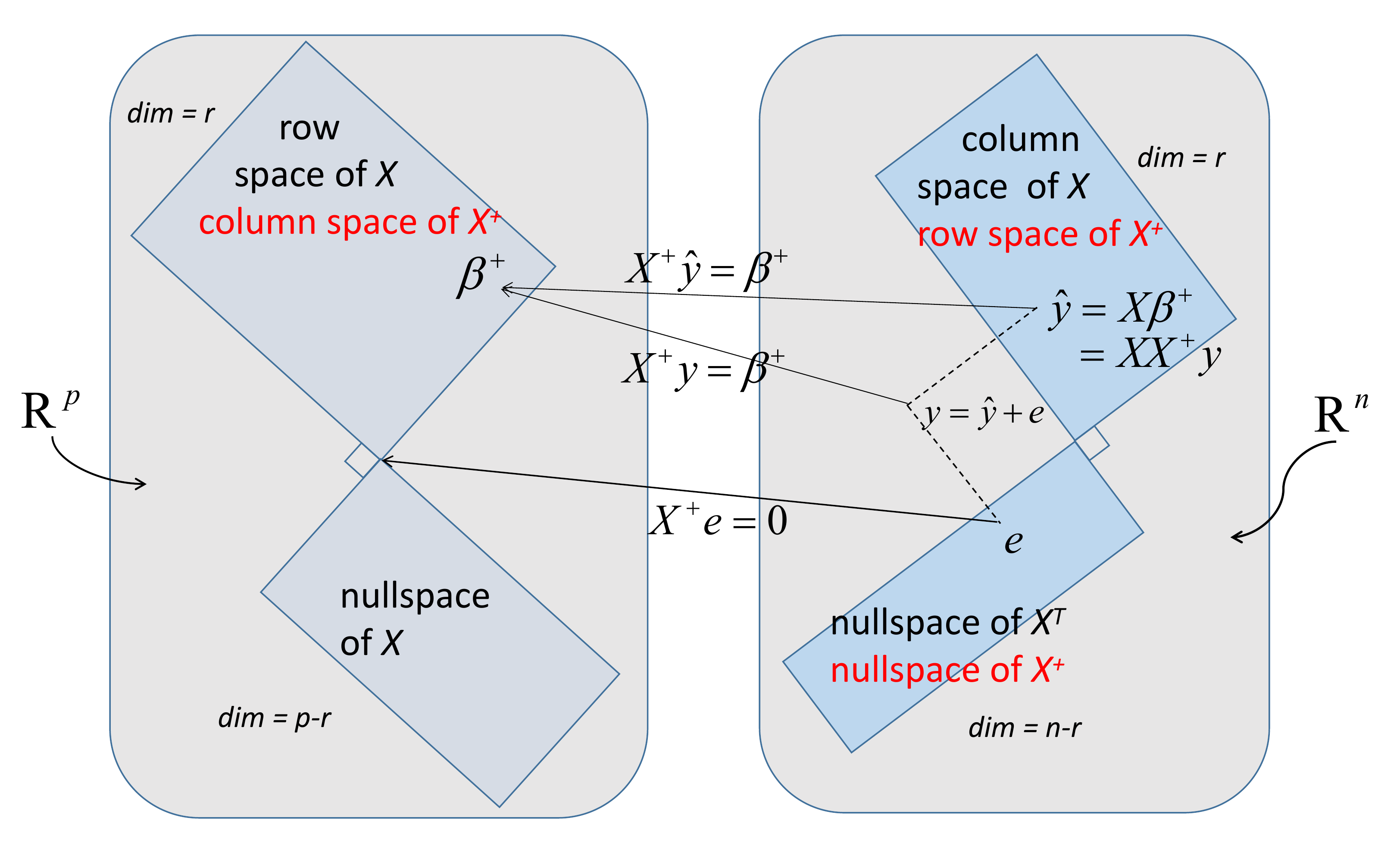}
	\caption{\textbf{Seventh Figure}: Column space and row space of pseudo-inverse $\bX^+$. $\bX$ transfers from row space to column space. $\bX^+$ transfers from  column space to row space. The split of $\by$ into $\hat{\by}+\be$ and the transformation to $\bbeta^+$ are discussed in Section~\ref{section:pseudo-in-svd}. \textbf{This is a more detailed picture of the pseudo-inverse compared to Figure~\ref{fig:lafundamental4-LS-SVD}}.}
	\label{fig:lafundamental5-pseudo}
\end{figure}

As we mentioned previously, for a matrix $\bX\in \real^{n\times p}$, we can find its pseudo-inverse, a $p\times n$ matrix denoted by $\bX^+$.
In words, when $\bX$ multiplies a vector $\bbeta$ in its row space, this produces $\bX\bbeta$ in the column space (see Figure~\ref{fig:lafundamental-ls}). Those two spaces have equal dimension $r$, i.e., the rank. $\bX$ is always invertible when restricted to these spaces and $\bX^+$ inverts $\bX$. That is, $\bX^+\bX\bbeta = \bbeta$ when $\bbeta$ is in the row space of $\bX$. And $\bX\bX^+\by = \by$ when $\by$ is in the column space of $\bX$ (see Figure~\ref{fig:lafundamental5-pseudo}).

The null space of $\bX^+$ is the null space of $\bX^\top$. It contains the vectors $\by$ in $\real^n$ with $\bX^\top\by = \bzero$. Those vectors $\by$ are perpendicular to every $\bX\bbeta$ in the column space. We delay the proof of this property in Lemma~\ref{lemma:pseudo-four-basis-space}. 

More formally, the pseudo-inverse, or also known as Moore-Penrose pseudo-inverse, $\bX^+$, is defined by the unique $p\times n$ matrix satisfying the following four criteria

\begin{equation}\label{equation:pseudi-four-equations}
	\boxed{
		\begin{aligned}
			&(C1) \qquad  \bX\bX^+\bX &=& \bX \qquad &(\bX^+\text{ is a g-inverse of }\bX) \\
			&(C2) \qquad  \bX^+\bX\bX^+ &=&\bX^+ \qquad &(\bX\text{ is a g-inverse of }\bX^+)\\
			&(C3) \qquad  (\bX\bX^+)^\top &=&\bX\bX^+\\
			&(C4) \qquad  	(\bX^+\bX)^\top &=& \bX^+\bX
		\end{aligned}
	}
\end{equation}

In Lemma~\ref{lemma:idempotent-of-ginverse}, we claimed that $\bX\bX^+$ and $\bX^+\bX$ are idempotent if $\bX^+$ is a g-inverse of $\bX$, and thus they are both projection matrix. For $\bX^+$ to be pseudo-inverse, by $(C3), (C4)$ conditions, they are symmetric such that they are orthogonal projection as well (again, please refer to \citep{lu2021rigorous} for more details about orthogonal projection matrices).

From the pseudo-inverse of the matrix from CR decomposition, we can also claim that any matrix has a pseudo-inverse.
\begin{svgraybox}
	\begin{lemma}[Existence of Pseudo-Inverse]\label{lemma:existence-of-pseudo-inverse}
		Every matrix $\bX$ has a pseudo-inverse.
	\end{lemma}
\end{svgraybox}
\begin{proof}[of Lemma~\ref{lemma:existence-of-pseudo-inverse}] For the CR decomposition of $\bX=\bC\bR$.
	\footnote{CR decomposition: Any rank-$r$ matrix $\bA \in \real^{n \times p}$ can be factored as $\bA = \bC \bR$, where $\bC$ is some $r$ independent columns of $\bA$, and $\bR$ is a $r\times p$ matrix to reconstruct the columns of $\bA$ from columns of $\bC$. In particular, $\bR$ is the row reduced echelon form of $\bA$ without the zero rows, and both $\bC$ and $\bR$ have full rank $r$.} Let 
	$$
	\bX^+ = \bR^+\bC^+ = \bR^\top (\bR\bR^\top)^{-1} (\bC^\top\bC)^{-1}\bC^\top,
	$$
	where $\bR^+=\bR^\top (\bR\bR^\top)^{-1}$ and $\bC^+ =(\bC^\top\bC)^{-1}\bC^\top$. \footnote{It can be easily checked that $\bR^+$ is the pseudo-inverse of $\bR$ and $\bC^+$ is the pseudo-inverse of $\bC$.} $\bR\bR^\top$ and $\bC^\top\bC$ are invertible since $\bC\in \real^{n\times r}$ and $\bR\in \real^{r\times p}$ have full rank $r$ from the property of CR decomposition. 
	
	Then, we can check that 
	$$
	\begin{aligned}
		&(C1) \qquad \bX\bX^+\bX &=& \bC\bR\left(\bR^\top (\bR\bR^\top)^{-1} (\bC^\top\bC)^{-1}\bC^\top\right)\bC\bR = \bC\bR = \bX, \\
		&(C2) \qquad \bX^+\bX\bX^+ &=&\left(\bR^\top (\bR\bR^\top)^{-1} (\bC^\top\bC)^{-1}\bC^\top\right) \bC\bR\left(\bR^\top (\bR\bR^\top)^{-1} (\bC^\top\bC)^{-1}\bC^\top\right)\\
		&&=&\bR^\top(\bR\bR^\top)^{-1}(\bC^\top\bC)^{-1}\bC^\top = \bX^+,\\
		&(C3) \qquad (\bX\bX^+)^\top &=& \bC(\bC^\top\bC)^{-1} (\bR\bR^\top)^{-1} \bR\bR^\top\bC^\top = \bC(\bC^\top\bC)^{-1} \bC^\top\\
		&&=&\bC\bR\bR^\top (\bR\bR^\top)^{-1} (\bC^\top\bC)^{-1}\bC^\top = \bX\bX^+,\\
		&(C4) \qquad (\bX^+\bX)^\top &=& \bR^\top\bC^\top \bC(\bC^\top\bC)^{-1} (\bR\bR^\top)^{-1} \bR= \bR^\top (\bR\bR^\top)^{-1} \bR\\
		&&=& \bR^\top (\bR\bR^\top)^{-1} (\bC^\top\bC)^{-1}\bC^\top\bC\bR = \bX^+\bX.
	\end{aligned}
	$$
	This implies $\bX^+$ is the pseudo-inverse of $\bX$ and the existence of the pseudo-inverse.
\end{proof}

\begin{svgraybox}
	\begin{lemma}[Uniqueness of Pseudo-Inverse]\label{lemma:uniqueness-of-pseudo-inverse}
		Every matrix $\bX$ has a unique pseudo-inverse.
	\end{lemma}
\end{svgraybox}
\begin{proof}[of Lemma~\ref{lemma:uniqueness-of-pseudo-inverse}]
	Suppose $\bX_1^+$ and $\bX_2^+$ are two pseudo-inverses of $\bX$. Then
	$$
	\begin{aligned}
		\bX_1^+ &= \bX_1^+\bX\bX_1^+ = \bX_1^+(\bX\bX_2^+\bX)\bX_1^+ = \bX_1^+ (\bX\bX_2^+)(\bX\bX_1^+) \qquad &(\text{by $(C2), (C1)$})\\
		&=\bX_1^+ (\bX\bX_2^+)^\top(\bX\bX_1^+)^\top =  \bX_1^+\bX_2^{+\top}\bX^\top \bX_1^{+\top}\bX^\top \qquad &(\text{by $(C3)$})\\
		&=\bX_1^+\bX_2^{+\top}(\bX \bX_1^{+}\bX)^\top= \bX_1^+\bX_2^{+\top}\bX^\top \qquad &(\text{by $(C1)$})\\
		&= \bX_1^+(\bX \bX_2^{+})^\top =\bX_1^+ \bX \bX_2^{+} = \bX_1^+ (\bX \bX_2^+\bX) \bX_2^{+}  \qquad &(\text{by $(C3),(C1)$})\\
		&= (\bX_1^+ \bX) (\bX_2^+\bX) \bX_2^{+} =  (\bX_1^+ \bX)^\top (\bX_2^+\bX)^\top \bX_2^{+}  \qquad &(\text{by $(C4)$})\\
		&=(\bX \bX_1^{+} \bX)^\top \bX_2^{+\top} \bX_2^{+} = \bX^\top \bX_2^{+\top} \bX_2^{+}   \qquad &(\text{by $(C1)$})\\
		&=  (\bX_2^{+}\bX)^\top \bX_2^{+} =\bX_2^{+}\bX \bX_2^{+} = \bX_2^{+}, \qquad &(\text{by $(C4),(C2)$})\\
	\end{aligned}
	$$
	which implies the uniqueness of pseudo-inverse.
\end{proof}

We are now ready to show the four subspaces in pseudo-inverse.
\begin{svgraybox}
	\begin{lemma}[Four Subspaces in Pseudo-Inverse]\label{lemma:pseudo-four-basis-space} For the pseudo-inverse $\bX^+$ of $\bX$, we have the following properties:
		
		$\bullet$ The column space of $\bX^+$ is the same as the row space of $\bX$;
		
		$\bullet$ The row space of $\bX^+$ is the same as the column space of $\bX$; 
		
		$\bullet$ The null space of $\bX^+$ is the same as the null space of $\bX^\top$; 
		
		$\bullet$ The null space of $\bX^{+\top}$ is the same as the null space of $\bX$.
		
		\item The relationship of the four subspaces is shown in Figure~\ref{fig:lafundamental5-pseudo}.
	\end{lemma}
\end{svgraybox}
\begin{proof}[of Lemma~\ref{lemma:pseudo-four-basis-space}]
Since $\bX^+$ is a special rg-inverse, by Lemma~\ref{Lemma:refle-ginverse-properties}, we have 
	$$
	\begin{aligned}
		\cspace(\bX\bX^+) &= \cspace(\bX) \qquad \text{and} \qquad \nspace(\bX^+\bX) = \nspace(\bX)\\
		\cspace(\bX^+\bX) &= \cspace(\bX^+)\qquad \text{and} \qquad \nspace(\bX\bX^+) = \nspace(\bX^+).
	\end{aligned}
	$$
	By $(C3)$ and $(C4)$ of the definition of pseudo-inverse, we also have 
	$$
	(\bX^+\bX)^\top = \bX^+\bX \qquad \text{and} \qquad (\bX\bX^+)^\top =\bX\bX^+.
	$$
	By fundamental theorem of linear algebra, we realize that $\cspace(\bX\bX^+)$ is the orthogonal complement to $\nspace((\bX\bX^+)^\top)$, and $\cspace(\bX^+\bX)$ is the orthogonal complement to $\nspace((\bX^+\bX)^\top)$: 
	$$
	\begin{aligned}
		\cspace(\bX\bX^+) \perp \nspace((\bX\bX^+)^\top) &\leadto \cspace(\bX\bX^+) \perp \nspace(\bX\bX^+) \\
		\cspace(\bX^+\bX) \perp \nspace((\bX^+\bX)^\top) &\leadto \cspace(\bX^+\bX) \perp \nspace(\bX^+\bX).
	\end{aligned}
	$$
	This implies 
	$$
	\begin{aligned}
		\cspace(\bX) \perp \nspace(\bX^+) \qquad \text{and} \qquad \cspace(\bX^+)\perp \nspace(\bX).
	\end{aligned}
	$$
	That is, $\nspace(\bX^+) = \nspace(\bX^\top)$ and $\cspace(\bX^+) = \cspace(\bX^\top)$. By fundamental theorem of linear algebra, this also implies $\cspace(\bX^{+\top})=\cspace(\bX)$ and $\nspace(\bX^{+\top})=\nspace(\bX)$.
\end{proof}

To conclude, we compare the properties for different inverses of $\bX$ in Table~\ref{table:different-inverses}.
\begin{table}[h!]
	\begin{tabular}{c|l|l|l}
		\hline
		& \multicolumn{1}{c|}{g-inverse}                                                                                   & \multicolumn{1}{c|}{rg-inverse}                                                                                                                                                                             & pseudo-inverse                                                                                                                                                                                                                                                                \\ \hline
		subspaces & \begin{tabular}[c]{@{}l@{}}$\cspace(\bX\bX^-) = \cspace(\bX)$\\ $\nspace(\bX^-\bX) = \nspace(\bX)$\end{tabular}  & \begin{tabular}[c]{@{}l@{}}$\cspace(\bX\bX_r^-) = \cspace(\bX)$ \\ $\nspace(\bX_r^-\bX) = \nspace(\bX)$\\ $\cspace(\bX_r^-\bX) = \cspace(\bX_r^-)$ \\ $\nspace(\bX\bX_r^-) = \nspace(\bX_r^-)$\end{tabular} & \begin{tabular}[c]{@{}l@{}}$\cspace(\bX\bX^+) = \cspace(\bX)=\cspace(\bX^{+\top})$ \\ $\nspace(\bX^+\bX) = \nspace(\bX)=\nspace(\bX^{+\top})$\\ $\cspace(\bX^+\bX) = \cspace(\bX^+)=\cspace(\bX^\top)$ \\ $\nspace(\bX\bX^+) = \nspace(\bX^+)=\nspace(\bX^\top)$\end{tabular} \\ \hline
		rank      & \begin{tabular}[c]{@{}l@{}}$rank(\bX\bX^-)$\\ $=rank(\bX^-\bX)$\\ $=rank(\bX)$\\ $\leq rank(\bX^-)$\end{tabular} & \begin{tabular}[c]{@{}l@{}}$rank(\bX_r^-)$\\ $=rank(\bX)$\\ $=rank(\bX\bX_r^-)$\\ $=rank(\bX_r^-\bX)$\end{tabular}                                                                                          & \begin{tabular}[c]{@{}l@{}}$rank(\bX^+)$\\ $=rank(\bX)$\\ $=rank(\bX\bX^+)$\\ $=rank(\bX^+\bX)$\end{tabular}                                                                                                                                                                  \\ \hline
	\end{tabular}\caption{Comparison of different inverses}\label{table:different-inverses}
\end{table}

\begin{svgraybox}
	\begin{lemma}[Projection onto Column Space and Row Space]\label{lemma:orthogonal-from-pseudo-inverse}
		For any matrix $\bX \in \real^{n\times p}$ and its pseudo-inverse $\bX^+ \in \real^{p\times n}$, $\bH = \bX\bX^+$ is the orthogonal projection onto column space of $\bX$. Similarly, $\bP= \bX^+\bX$ is the orthogonal projection onto row space of $\bX$.
	\end{lemma}
\end{svgraybox}
\begin{proof}[of Lemma~\ref{lemma:orthogonal-from-pseudo-inverse}]
	As $\bH^\top = (\bX\bX^+)^\top=\bX\bX^+=\bH$ from the definition of the pseudo-inverse, and $\bH$ is idempotent such that $\bH$ is an orthogonal projection.
	From Table~\ref{table:different-inverses}, we conclude that $\cspace(\bH)=\cspace(\bX\bX^+)=\cspace(\bX)$. This implies $\bH$ is the orthogonal projection onto the column space of $\bX$. Similarly, we can prove $\bP= \bX^+\bX$ is the orthogonal projection onto row space of $\bX$.
\end{proof}

\begin{mdframed}[hidealllines=\mdframehidelineNote,backgroundcolor=\mdframecolorNote,frametitle={Pseudo-Inverse in Different Cases}]
	Specifically, we define in either of the following ways:
	
	$\bullet$ Case $n>p=r$, i.e., matrix $\bX$ has independent columns: 
	
	$\bX^\top\bX$ is a $p\times p$ invertible matrix, and we define the left-pseudo-inverse:
	$$
	\boxed{\text{left-pseudo-inverse} = \bX^+ = (\bX^\top\bX)^{-1}\bX^\top},
	$$
	which satisfies
	$$
	\bX^+ \bX =  (\bX^\top\bX)^{-1}\bX^\top\bX = \bI_p.
	$$
	But 
	$$
	\bX\bX^+ = \bX(\bX^\top\bX)^{-1}\bX^\top \neq\bI.
	$$
	
	$\bullet$ Case $p>n=r$, i.e., matrix $\bX$ has independent rows: 
	
	$\bX\bX^\top$ is an $n\times n$ invertible matrix, and we define the right-pseudo-inverse:
	$$
	\boxed{\text{right-pseudo-inverse} = \bX^+ = \bX^\top(\bX\bX^\top)^{-1}},
	$$
	which satisfies 
	$$
	\bX\bX^+ = \bX\bX^\top(\bX\bX^\top)^{-1} = \bI_n.
	$$
	But 
	\begin{equation}\label{equation:right-inverse-ba-appendix}
		\bX^+ \bX =  \bX^\top(\bX\bX^\top)^{-1}\bX \neq \bI.
	\end{equation}
	
	$\bullet$ Case rank-deficient: we delay the pseudo-inverse for rank-deficient matrices in the next section.
	
	We can also show that $(\bX^+)^+ = \bX$. If $n>p=r$, we have 
	$$
	\begin{aligned}
		(\bX^+)^+ &= [(\bX^\top\bX)^{-1}\bX^\top]^+ \\
		&= \bX^{+\top}(\bX^+\bX^{+\top})^{-1} &(n>p=r) \\
		&= \left[(\bX^\top\bX)^{-1}\bX^\top\right]^{\top}\left\{\left[(\bX^\top\bX)^{-1}\bX^\top\right]\left[(\bX^\top\bX)^{-1}\bX^\top\right]^{\top}\right\}^{-1}\\
		&=\bX (\bX^\top\bX)^{-1}\left\{(\bX^\top\bX)^{-1}\bX^\top\bX(\bX^\top\bX)^{-1}   \right\}^{-1}\\
		&=\bX.
	\end{aligned}
	$$
	Similarly, we can show $(\bX^+)^+ = \bX$ if $p>n=r$.
	
	In particular, when $n=p$, $\bX$ is square invertible matrix, then both left and right-pseudo-inverse are the inverse of $\bX$:
	$$
	\begin{aligned}
		\text{left-pseudo-inverse} &= \bX^+ = (\bX^\top\bX)^{-1}\bX^\top=\bX^{-1} \bX^{-\top} \bX^\top = \bX^{-1}, \\
		\text{right-pseudo-inverse} &= \bX^+ = \bX^\top(\bX\bX^\top)^{-1} = \bX^\top\bX^{-\top}\bX^{-1} = \bX^{-1}.
	\end{aligned}
	$$
\end{mdframed}

\subsection{Pseudo-Inverse in SVD}\label{section:pseudo-in-svd}
\begin{mdframed}[hidealllines=\mdframehidelineNote,backgroundcolor=\mdframecolorNote,frametitle={Pseudo-Inverse in Different Cases via SVD}]
	For full SVD of matrix $\bX = \bU\bSigma\bV^\top$. Consider the following cases:
	
	$\bullet$ Case $n>p=r$: 
	$$
	\begin{aligned}
		\text{left-pseudo-inverse} &= \bX^+ = (\bX^\top\bX)^{-1}\bX^\top \\
		&= (\bV \bSigma^\top \bU^\top\bU\bSigma\bV^\top)^{-1}\bV \bSigma^\top \bU^\top\\
		&= \bV (\bSigma^\top\bSigma)^{-1}\bV^\top \bV \bSigma^\top \bU^\top \\
		&= \bV[(\bSigma^\top\bSigma)^{-1}\bSigma^\top] \bU^\top \\
		&= \bV \bSigma^+ \bU^\top.  \qquad &(\bSigma^+=(\bSigma^\top\bSigma)^{-1}\bSigma^\top)
	\end{aligned}
	$$
	
	$\bullet$ Case $p>n=r$: 
	$$
	\begin{aligned}
		\text{right-pseudo-inverse} &= \bX^+ = \bX^\top(\bX\bX^\top)^{-1}\\
		&= (\bU\bSigma\bV^\top)^\top[(\bU\bSigma\bV^\top)(\bU\bSigma\bV^\top)^\top]^{-1}\\
		&= \bV\bSigma^\top\bU^\top(\bU\bSigma\bV^\top\bV\bSigma^\top\bU^\top)^{-1} \\
		&= \bV\bSigma^\top\bU^\top\bU^{-\top}(\bSigma\bSigma^\top)^{-1}\bU^{-1} \\
		&= \bV\bSigma^\top(\bSigma\bSigma^\top)^{-1}\bU^{-1} \\
		&=\bV \bSigma^+ \bU^\top.  \qquad &(\bSigma^+=\bSigma^\top(\bSigma\bSigma^\top)^{-1})
	\end{aligned}
	$$
	
	$\bullet$ Case rank-deficient: $\bX^+=\bV \bSigma^+ \bU^\top$,
	where the upper-left side of $\bSigma^+ \in \real^{p\times n}$ is a diagonal matrix $diag(\frac{1}{\sigma_1}, \frac{1}{\sigma_2}, \cdots, \frac{1}{\sigma_r})$. It can be easily verified that this definition of pseudo-inverse satisfies the four criteria in Equation~\ref{equation:pseudi-four-equations}.
\end{mdframed}

In either case, we have $\bSigma^+$ as the pseudo-inverse of $\bSigma$ with $1/\sigma_1, 1/\sigma_2, \cdots, 1/\sigma_r$ on its diagonal. We thus conclude the pseudo-inverse from SVD in the Table~\ref{table:pseudo-inverse-svd}.
\begin{table}[h!]\centering
	\begin{tabular}{c|c|c|c|c}
		\hline
		& $\bX$                & $\bX^\top$           & $\bX^+$                & $\bA^{+\top}$          \\ \hline
		SVD & $\bU\bSigma\bV^\top$ & $\bV\bSigma\bU^\top$ & $\bV\bSigma^+\bU^\top$ & $\bU\bSigma^+\bV^\top$ \\ \hline
	\end{tabular}\caption{Pseudo-inverse in SVD}\label{table:pseudo-inverse-svd}
\end{table}

From the pseudo-inverse via SVD, we can provide another way to see the orthogonal projection in pseudo-inverse.
\begin{mdframed}[hidealllines=\mdframehidelineNote,backgroundcolor=\mdframecolorNote,frametitle={Another Way to See the Orthogonal Projection in Pseudo-Inverse via SVD}]
	
	We have shown previously that $\bH=\bX\bX^+$ is an orthogonal projection, so we only need to show that it projects onto $\cspace(\bX)$. For any vector $\by\in \real^n$, we have 
	$$
	\bH\by= \bX\bX^+\by = \bX\bbeta^+,
	$$
	which is a linear combination of columns of $\bX$. Thus $\cspace(\bH)\subseteq \cspace(\bX)$.
	
	Moreover, since $\bH$ is a symmetric idempotent matrix, by Lemma~\ref{lemma:rank-of-symmetric-idempotent2}, we have $rank(\bH)=trace(\bH)=trace(\bX\bX^+) = trace(\bU\bSigma\bV^\top \bV\bSigma^+\bU^\top)=trace(\bU\bSigma\bSigma^+\bU^\top)=r$, where $\bU\bSigma\bV^\top$ is the SVD of $\bX$. Then $\cspace(\bH)= \cspace(\bX)$ and we complete the proof.
	
	Similarly, we can prove $\bP= \bX^+\bX$ is the orthogonal projection onto the row space of $\bX$.
\end{mdframed}

We finally prove the important property of the four subspaces in the pseudo-inverse $\bX^+$ via SVD. 
Firstly, we need to show the following lemma that $\bX^+ \bX^{+\top}$ and $\bX^+$ have same rank.
\begin{svgraybox}
	\begin{lemma}[Rank of $\bX^+ \bX^{+\top}$]\label{lemma:apapt-rank}
		$\bX^+ \bX^{+\top}$ and $\bX^+$ have same rank.
	\end{lemma}
\end{svgraybox}
The claim in the lemma above and the proof are similar to Lemma~\ref{lemma:rank-of-ata-x}.
\begin{proof}[of Lemma~\ref{lemma:apapt-rank}]
	Let $\bbeta\in \nspace(\bX^{+\top})$, where $\bX^{+\top}$ is the transpose of $\bX^{+}$, we have 
	$$
	\bX^{+\top}\bbeta  = \bzero \leadto \bX^+\bX^{+\top} \bbeta =\bzero, 
	$$
	i.e., $\bbeta\in \nspace(\bX^{+\top}) \leadtosmall \bbeta \in \nspace(\bX^+\bX^{+\top})$, therefore $\nspace(\bX^{+\top}) \in \nspace(\bX^+\bX^{+\top})$. 
	
	Further, let $\bbeta \in \nspace(\bX^+\bX^{+\top})$, we have 
	$$
	\bX^+\bX^{+\top}\bbeta = \bzero\leadtosmall \bbeta^\top \bX^+\bX^{+\top}\bbeta = 0\leadtosmall ||\bX^{+\top}\bbeta||^2 = 0 \leadtosmall \bX^{+\top}\bbeta=\bzero, 
	$$
	i.e., $\bbeta\in \nspace(\bX^+\bX^{+\top}) \leadtosmall \bbeta\in \nspace(\bX^{+\top}\bbeta)$, therefore $\nspace(\bX^+\bX^{+\top}) \in\nspace(\bX^{+\top}) $. As a result, $\nspace(\bX^{+\top}) = \nspace(\bX^+\bX^{+\top})$ and $dim(\nspace(\bX^{+\top})) = dim(\nspace(\bX^+\bX^{+\top}))$. By fundamental theorem of linear algebra in Section~\ref{appendix:fundamental-rank-nullity}, $\bX^+ \bX^{+\top}$ and $\bX^+$ have same rank.
\end{proof}
By the lemma above, we can provide another way to show the four subspaces in pseudo-inverse.
\begin{mdframed}[hidealllines=\mdframehidelineNote,backgroundcolor=\mdframecolorNote,frametitle={Another Way to See the Subspaces in Pseudo-Inverse via SVD}]
	From Lemma~\ref{lemma:rank-of-symmetric}, for symmetric matrix $\bX^+ \bX^{+\top} = \bV(\bSigma^\top\bSigma)^{-1}\bV^\top$, $\cspace(\bX^+ \bX^{+\top})$ is spanned by the eigenvectors, thus $\{\bv_1,\bv_2 \cdots, \bv_r\}$ is an orthonormal basis of $\cspace(\bX^+ \bX^{+\top})$.
	
	Since, 
	
	1. $\bX^+ \bX^{+\top}$ is symmetric, then the row space of $\bX^+ \bX^{+\top}$ equals to the column space of $\bX^+ \bX^{+\top}$. 
	
	2. All columns of $\bX^+ \bX^{+\top}$ are combination of columns of $\bX^+$, so the column space of $\bX^+ \bX^{+\top}$ $\subseteq$ column space of $\bX^+$, i.e., $\cspace(\bX^+ \bX^{+\top}) \subseteq \cspace(\bX^+)$. 
	
	3. Since $rank(\bX^+ \bX^{+\top}) = rank(\bX^+)$ by Lemma~\ref{lemma:apapt-rank}, we then have 
	
	The row space of $\bX^+ \bX^{+\top}$ = the column space of $\bX^+ \bX^{+\top}$ =  the column space of $\bX^+$, i.e., $\cspace(\bX^+ \bX^{+\top}) = \cspace(\bX^+ )$. \textcolor{blue}{Thus $\{\bv_1, \bv_2,\cdots, \bv_r\}$ is an orthonormal basis of $\cspace(\bX^+)$}. We also proved in Lemma~\ref{lemma:svd-four-orthonormal-Basis} that $\{\bv_1, \bv_2,\cdots, \bv_r\}$ is an orthonormal basis of the row space of $\bX$ (i.e., basis of $\cspace(\bX^\top)$). So $\cspace(\bX^+)=\cspace(\bX^\top)$ as shown in Figure~\ref{fig:lafundamental5-pseudo}.
	
	Similarly, if we apply this process to $ \bX^{+\top}\bX^+$, we can show the row space of $\bX^+$ is equal to the column space of $\bX$, and the null space of $\bX^+$ is equal to the null space of $\bX^\top$.
	%
\end{mdframed}

Moreover, for $\bbeta^+$ in row space of $\bX$, we have $\bbeta^+ = \sum_{i=1}^{r} x_i \bv_i$ since $\{\bv_1, \bv_2, \cdots, \bv_r\}$ is an orthonormal basis for the row space of $\bX$. For vector $\hat{\by} = \bX\bbeta^+$ in the column space of $\bX$, we have $\hat{\by} = \bU\bSigma\bV^\top \bbeta^+$ and 
$$
\bX^+\hat{\by} = \bV\bSigma^+\bU^\top \bU\bSigma\bV^\top \bbeta^+ = \bV\bSigma^+\bSigma\bV^\top \bbeta^+ =(\sum_{i=1}^{r} \bv_i \bv_i^\top) (\sum_{i=1}^{r} x_i \bv_i) = \sum_{i=1}^{r} x_i \bv_i = \bbeta^+.
$$
In conclude, for any vector $\bbeta^+$ in row space of $\bX$, we have 
$$
\hat{\by} = \bX\bbeta^+ \leadto \bX^+\hat{\by}=\bbeta^+,
$$
and the relationship is depicted in Figure~\ref{fig:lafundamental5-pseudo}.

%
%




\newpage
\appendix

\section{Rank of a Special Symmetric Matrix}

\begin{svgraybox}
	\begin{lemma}[Rank of $\bX^\top \bX$]\label{lemma:rank-of-ata-x} 
		For any matrix $\bX$, $\bX^\top \bX$ and $\bX$ have same rank.
	\end{lemma}
\end{svgraybox}
\begin{proof}[of Lemma~\ref{lemma:rank-of-ata-x}]
	Let $\boldeta\in \nspace(\bX)$, we have 
	$$
	\bX\boldeta  = \bzero \leadto \bX^\top\bX \boldeta =\bzero, 
	$$
	i.e., $\boldeta\in \nspace(\bX) \leadtosmall \boldeta \in \nspace(\bX^\top \bX)$. Therefore, $\nspace(\bX) \in \nspace(\bX^\top\bX)$. 
	
	Further, let $\boldeta \in \nspace(\bX^\top\bX)$, we have 
	$$
	\bX^\top \bX\boldeta = \bzero\leadtosmall \boldeta^\top \bX^\top \bX\boldeta = 0\leadtosmall ||\bX\boldeta||^2 = 0 \leadtosmall \bX\boldeta=\bzero, 
	$$
	i.e., $\boldeta\in \nspace(\bX^\top \bX) \leadtosmall \boldeta\in \nspace(\bX)$. Therefore, $\nspace(\bX^\top\bX) \in\nspace(\bX) $. As a result, $\nspace(\bX) = \nspace(\bX^\top\bX)$ and $dim(\nspace(\bX)) = dim(\nspace(\bX^\top\bX))$. By fundamental theorem of linear algebra in Section~\ref{appendix:fundamental-rank-nullity}, $\bX^\top \bX$ and $\bX$ have same rank.
\end{proof}

\section{Similar Matrices}\label{appendix:similar-matrices}
\begin{svgraybox}
	\begin{definition}[Similar Matrices]
		For any nonsingular matrix $\bP$, the matrices $\bA$ and $\bP\bA\bP^{-1}$ are called similar matrices.
	\end{definition}
\end{svgraybox}
\begin{svgraybox}
	\begin{lemma}[Eigenvalue and Rank of Similar Matrices]\label{lemma:eigenvalue-similar-matrices}
		Any eigenvalue of $\bA$ is also an eigenvalue of $\bP\bA\bP^{-1}$ for any nonsingular matrix $\bP$. The converse is also true that any eigenvalue of $\bP\bA\bP^{-1}$ is also an eigenvalue of $\bA$.
		
		And also the rank of $\bA$ is equal to the rank of matrix $\bP\bA\bP^{-1}$ for any nonsingular matrix $\bP$.
	\end{lemma}
\end{svgraybox}
\begin{proof}[of Lemma~\ref{lemma:eigenvalue-similar-matrices}]
	For any eigenvalue $\lambda$ of $\bA$, we have $\bA\bx =\lambda \bx$. Then $\lambda \bP\bx = \bP\bA\bP^{-1} \bP\bx$ such that $\bP\bx$ is an eigenvector of $\bP\bA\bP^{-1}$ corresponding to $\lambda$.
	
	Similarly, for any eigenvalue $\lambda$ of $\bP\bA\bP^{-1}$, we have $\bP\bA\bP^{-1} \bx = \lambda \bx$. Then $\bA\bP^{-1} \bx = \lambda \bP^{-1}\bx$ such that $\bP^{-1}\bx$ is an eigenvector of $\bA$ corresponding to $\lambda$. 
	
	For the rank of $\bP\bA\bP^{-1}$, we have $trace(\bP\bA\bP^{-1}) = trace(\bA\bP^{-1}\bP) = trace(\bA)$, where the first equality comes from the fact that 
	trace of a product is invariant under cyclical permutations of the factors:
	\begin{equation}
		\boxed{trace(\bA\bB\bC) = trace(\bB\bC\bA) = trace(\bC\bA\bB)}, \nonumber
	\end{equation}
	if all $\bA\bB\bC$, $\bB\bC\bA$, and $\bC\bA\bB$ exist.
\end{proof}

\section{Properties of Symmetric and Idempotent Matrices}

Symmetric idempotent matrices have specific eigenvalues which will be often used.
\begin{svgraybox}
	\begin{lemma}[Eigenvalue of Symmetric Idempotent Matrices]\label{proposition:eigenvalues-of-projection}
		The only possible eigenvalues of any symmetric idempotent matrix are 0 and 1.
	\end{lemma}
\end{svgraybox}
In Lemma~\ref{proposition:eigenvalues-of-projection2}, we prove the eigenvalues of idempotent matrices are 1 and 0 as well which relaxes the conditions required here (both idempotent and symmetric). However, the method used in the proof is quite useful so we keep both of the claims. 
\begin{proof}[of Lemma~\ref{proposition:eigenvalues-of-projection}]
	Suppose matrix $\bA$ is symmetric idempotent. By spectral theorem (Theorem~\ref{theorem:spectral_theorem}), we can decompose $\bA = \bQ \bLambda\bQ^\top$, where $\bQ$ is an orthogonal matrix and $\bLambda$ is a diagonal matrix. Therefore,
	\begin{equation}
		\begin{aligned}
			(\bQ\bLambda\bQ^\top)^2 &= \bQ\bLambda\bQ \\
			\bQ\bLambda\bQ^\top\bQ\bLambda\bQ^\top &= \bQ\bLambda\bQ^\top \\
			\bQ\bLambda^2\bQ^\top &= \bQ\bLambda\bQ^\top \\
			\bLambda^2 &= \bLambda \\
			\lambda_i^2 &= \lambda_i \nonumber.
		\end{aligned}
	\end{equation}
	Thus the eigenvalues of $\bA$ satisfies that $\lambda_i \in \{0,1\},\,\, \forall i$. We complete the proof.
\end{proof}

In above lemma, we use spectral theorem to prove the only eigenvalues of any symmetric idempotent matrices are 1 and 0. This trick from spectral theorem is often used in mathematical proofs (see distribution theory sections in \citep{lu2021rigorous}). By trivial trick, we can relax the condition from symmetric idempotent to idempotent.
\begin{svgraybox}
	\begin{lemma}[Eigenvalue of Idempotent Matrices]\label{proposition:eigenvalues-of-projection2}
		The only possible eigenvalues of any idempotent matrix are 0 and 1.
	\end{lemma}
\end{svgraybox}
\begin{proof}[of Lemma~\ref{proposition:eigenvalues-of-projection2}]
	Let $\bx$ be an eigenvector of the idempotent matrix $\bA$ corresponding to eigenvalue $\lambda$. That is
	$$
	\bA \bx = \lambda \bx.
	$$
	Also, we have 
	$$
	\begin{aligned}
		\bA^2\bx &=(\bA^2)\bx =\bA\bx = \lambda\bx\\
		&=\bA(\bA\bx) = \bA(\lambda\bx)=\lambda\bA\bx=\lambda^2\bx,
	\end{aligned}
	$$
	which implies $\lambda^2 = \lambda$ and $\lambda$ is 0 or 1.
\end{proof}

We also prove the rank of a symmetric idempotent matrix.
\begin{svgraybox}
	\begin{lemma}[Rank and Trace of Symmetric Idempotent Matrices]\label{lemma:rank-of-symmetric-idempotent}
		For any $n\times n$ symmetric idempotent matrix $\bA$, the rank of $\bA$ equals to the trace of $\bA$.
	\end{lemma}
\end{svgraybox}
\begin{proof}[of Lemma~\ref{lemma:rank-of-symmetric-idempotent}]
	From spectral theorem~\ref{theorem:spectral_theorem}, we have spectral decomposition for $\bA = \bQ\bLambda\bQ^\top$. 
	Since $\bA$ and $\bLambda$ are similar matrices, their rank and trace are the same (see Appendix~\ref{appendix:similar-matrices}). That is, 
	$$
	\begin{aligned}
		rank(\bA) &= rank(\diag(\lambda_1, \lambda_2, \cdots, \lambda_n)),\\
		trace(\bA) &= trace(\diag(\lambda_1, \lambda_2, \cdots, \lambda_n)),\\
	\end{aligned}
	$$
	By Lemma~\ref{proposition:eigenvalues-of-projection}, the only eigenvalues of $\bA$ are 0 and 1. Then, 
	$rank(\bA) = trace(\bA)$.
\end{proof}

In above lemma, we prove the rank and trace of any symmetric idempotent matrix are the same. However, it is rather a loose condition. We here also prove that the only condition on idempotency has same result. Again, although this lemma is a more general version, we provide both of them since the method used in the proof is quite useful.
\begin{svgraybox}
	\begin{lemma}[Rank and Trace of an Idempotent Matrix]\label{lemma:rank-of-symmetric-idempotent2}
		For any $n\times n$ idempotent matrix $\bA$, the rank of $\bA$ equals to the trace of $\bA$.
	\end{lemma}
\end{svgraybox}
\begin{proof}[of Lemma~\ref{lemma:rank-of-symmetric-idempotent2}]
	Any $n\times n$  rank-$r$ matrix $\bA$ has CR decomposition $\bA = \bC\bR$, where $\bC\in\real^{n\times r}$ and $\bR\in \real^{r\times n}$ with $\bC, \bR$ having full rank $r$ (see \citep{lu2021numerical}).\footnote{CR decomposition: Any rank-$r$ matrix $\bA \in \real^{n \times p}$ can be factored as $\bA = \bC \bR$, where $\bC$ is some $r$ independent columns of $\bA$, and $\bR$ is a $r\times p$ matrix to reconstruct the columns of $\bA$ from columns of $\bC$. In particular, $\bR$ is the row reduced echelon form of $\bA$ without the zero rows, and both $\bC$ and $\bR$ have full rank $r$.} 
	Then, 
	$$
	\begin{aligned}
		\bA^2 &= \bA \\
		\bC\bR\bC\bR &= \bC\bR \\
		\bR\bC\bR &=\bR \\
		\bR\bC &=\bI_r,
	\end{aligned}
	$$ 
	where $\bI_r$ is a $r\times r$ identity matrix. Thus
	$$
	trace(\bA) = trace(\bC\bR) =trace(\bR\bC) = trace(\bI_r) = r, 
	$$
	which equals to the rank of $\bA$.
\end{proof}

\section{Spectral Decomposition (Theorem)}\label{appendix:spectral-decomposition}

In this section, we provide the proof for Theorem~\ref{theorem:spectral_theorem} which is essential for the existence of singular value decomposition in the sequel.
\begin{svgraybox}
	\begin{theorem}[Spectral Decomposition]\label{theorem:spectral_theorem}
		A real matrix $\bA \in \real^{n\times n}$ is symmetric if and only if there exists an orthogonal matrix $\bQ$ and a diagonal matrix $\bLambda$ such that
		\begin{equation*}
			\bA = \bQ \bLambda \bQ^\top,
		\end{equation*}
		where the columns of $\bQ = [\bq_1, \bq_2, \cdots, \bq_n]$ are eigenvectors of $\bA$ and orthonormal, and the entries of $\bLambda=diag(\lambda_1, \lambda_2, \cdots, \lambda_n)$ are the corresponding eigenvalues of $\bA$, which are real. Specifically, we have the following properties:
		
		1. A symmetric matrix has only \textbf{real eigenvalues};
		
		2. The eigenvectors can be chosen \textbf{orthonormal};
		
		3. The rank of $\bA$ is the number of nonzero eigenvalues;
		
		4. If the eigenvalues are distinct, the eigenvectors are unique as well.
	\end{theorem}
\end{svgraybox}



We prove the theorem in several steps. 
\begin{tcolorbox}[title={Symmetric Matrix Property 1 of 4}]
	The eigenvalues of any symmetric matrix are all real. 
\end{tcolorbox}
\begin{proof}[\textbf{All the eigenvalues of a real symmetric matrix are real}]
	Suppose eigenvalue $\lambda$ is a complex number $\lambda=a+ib$ where $a,b$ are real. Its complex conjugate is $\bar{\lambda}=a-ib$. Same for complex eigenvector $\bbeta = \bc+i\bd$ and its complex conjugate $\bar{\bbeta}=\bc-i\bd$ where $\bc, \bd$ are real vectors. We then have the following property
	$$
	\bX \bbeta = \lambda \bbeta\qquad   \underrightarrow{\text{ leads to }}\qquad  \bX \bar{\bbeta} = \bar{\lambda} \bar{\bbeta}\qquad   \underrightarrow{\text{ transpose to }}\qquad  \bar{\bbeta}^\top \bX =\bar{\lambda} \bar{\bbeta}^\top.
	$$
	We take dot product of the first equation with $\bar{\bbeta}$ and the last equation with $\bbeta$:
	$$
	\bar{\bbeta}^\top \bX \bbeta = \lambda \bar{\bbeta}^\top \bbeta, \qquad \text{and } \qquad \bar{\bbeta}^\top \bX \bbeta = \bar{\lambda}\bar{\bbeta}^\top \bbeta.
	$$
	Then we have the equality $\lambda\bar{\bbeta}^\top \bbeta = \bar{\lambda} \bar{\bbeta}^\top\bbeta$. Since $\bar{\bbeta}^\top\bbeta = (\bc-i\bd)^\top(\bc+i\bd) = \bc^\top\bc+\bd^\top\bd$ is a real number. Therefore the imaginary part of $\lambda$ is zero and $\lambda$ is real.
\end{proof}

\begin{tcolorbox}[title={Symmetric Matrix Property 2 of 4}]
	The eigenvectors  corresponding to distinct eigenvalues of any symmetric matrix are orthogonal so that we can normalize eigenvectors to make them orthonormal since $\bX\bbeta = \lambda \bbeta \underrightarrow{\text{ leads to } } \bX\frac{\bbeta}{||\bbeta||} = \lambda \frac{\bbeta}{||\bbeta||}$ which corresponds to the same eigenvalue.
\end{tcolorbox}
\begin{proof}[\textbf{The eigenvectors are orthogonal}]
	Suppose eigenvalues $\lambda_1, \lambda_2$ correspond to eigenvectors $\bbeta_1, \bbeta_2$ so that $\bX\bbeta_1=\lambda \bbeta_1$ and $\bX\bbeta_2 = \lambda_2\bbeta_2$. We have the following equality:
	$$
	\bX\bbeta_1=\lambda_1 \bbeta_1 \leadto \bbeta_1^\top \bX =\lambda_1 \bbeta_1^\top \leadto \bbeta_1^\top \bX \bbeta_2 =\lambda_1 \bbeta_1^\top\bbeta_2,
	$$
	and 
	$$
	\bX\bbeta_2 = \lambda_2\bbeta_2 \leadto  \bbeta_1^\top\bX\bbeta_2 = \lambda_2\bbeta_1^\top\bbeta_2,
	$$
	which implies $\lambda_1 \bbeta_1^\top\bbeta_2=\lambda_2\bbeta_1^\top\bbeta_2$. Since eigenvalues $\lambda_1\neq \lambda_2$, the eigenvectors are orthogonal.
\end{proof}

For any matrix multiplication we have the rank of the multiplication result no larger than the rank of the inputs. However, the symmetric matrix $\bX^\top \bX$ is rather special that the rank of $\bX^\top \bX$ is equal to that of $\bX$ which will be used in the proof of singular value decomposition in the next section. 
\begin{svgraybox}
	\begin{lemma}[Rank of $\bX\bY$]\label{lemma:rankAB}
		For any matrix $\bX\in \real^{m\times n}$, $\bY\in \real^{n\times k}$, then the matrix multiplication $\bX\bY\in \real^{m\times k}$ has $rank$($\bX\bY$)$\leq$min($rank$($\bX$), $rank$($\bY$)).
	\end{lemma}
\end{svgraybox}

\begin{proof}[of Lemma~\ref{lemma:rankAB}]
	For matrix multiplication $\bX\bY$, we have 
	
	$\bullet$ All rows of $\bX\bY$ are combination of the rows of $\bY$, the row space of $\bX\bY$ is a subset of the row space of $\bY$. Thus $rank$($\bX\bY$)$\leq$$rank$($\bY$).
	
	$\bullet$ All columns of $\bX\bY$ are combination of columns of $\bX$, the column space of $\bX\bY$ is a subset of the column space of $\bX$. Thus $rank$($\bX\bY$)$\leq$$rank$($\bX$).
	
	Therefore we have, $rank$($\bX\bY$)$\leq$min($rank$($\bX$), $rank$($\bY$)).
\end{proof}

\begin{tcolorbox}[title={Symmetric Matrix Property 3 of 4}]
	\begin{lemma}[Rank of Symmetric Matrix]\label{lemma:rank-of-symmetric}
		If $\bX$ is an $n\times n$ real symmetric matrix, then rank($\bX$) =
		the total number of nonzero eigenvalues of $\bX$. 
		In particular, $\bX$ has full rank if and only if $\bX$ is nonsingular. Further, $\cspace(\bX)$ is the linear space spanned by the eigenvectors of $\bX$ that correspond to nonzero eigenvalues.
	\end{lemma}
\end{tcolorbox}
\begin{proof}[of Lemma~\ref{lemma:rank-of-symmetric}]
	For any symmetric matrix $\bX$, we have $\bX$, in spectral form, as $\bX = \bQ \bLambda\bQ^\top$ and also $\bLambda = \bQ^\top\bX\bQ$. Since we have shown in Lemma~\ref{lemma:rankAB} that the rank of the matrix multiplication $rank$($\bX\bY$)$\leq$min($rank$($\bX$), $rank$($\bY$)).
	
	$\bullet$ From $\bX = \bQ \bLambda\bQ^\top$, we have $rank(\bX) \leq rank(\bQ \bLambda) \leq rank(\bLambda)$;
	
	$\bullet$ From $\bLambda = \bQ^\top\bX\bQ$, we have $rank(\bLambda) \leq rank(\bQ^\top\bX) \leq rank(\bX)$, 
	
	The inequalities above give us contradiction. And thus $rank(\bX) = rank(\bLambda)$ which is the total number of nonzero eigenvalues.
	
	Since $\bX$ is nonsingular if and only if all of its eigenvalues are nonzero, $\bX$ has full rank if and only if $\bX$ is nonsingular.
\end{proof}

For the fourth property of symmetric matrix, we need the definition of similar matrices and the property about eigenvalues of similar matrices (see Appendix~\ref{appendix:similar-matrices}).
%
%

\begin{tcolorbox}[title={Symmetric Matrix Property 4 of 4}]
	\begin{lemma}[Orthonormal Eigenvectors for Duplicate Eigenvalue]\label{lemma:eigen-multiplicity}
		If $\bX$ has a duplicate eigenvalue $\lambda_i$ with multiplicity $k\geq 2$, then there exist $k$ orthonormal eigenvectors corresponding to $\lambda_i$.
	\end{lemma}
\end{tcolorbox}
\begin{proof}[of Lemma~\ref{lemma:eigen-multiplicity}]
	We note that there is at least one eigenvector $\bbeta_{i1}$ corresponding to $\lambda_i$. And for such eigenvector $\bbeta_{i1}$, we can always find additional $n-1$ orthonormal vectors $\by_2, \by_3, \cdots, \by_n$ so that $\{\bbeta_{i1}, \by_2, \by_3, \cdots, \by_n\}$ forms an orthonormal basis in $\real^n$. Put the $\by_2, \by_3, \cdots, \by_n$ into matrix $\bY_1$ and $\{\bbeta_{i1}, \by_2, \by_3, \cdots, \by_n\}$ into matrix $\bP_1$
	$$
	\bY_1=[\by_2, \by_3, \cdots, \by_n] \qquad \text{and} \qquad \bP_1=[\bbeta_{i1}, \bY_1].
	$$
	We then have
	$$
	\bP_1^\top\bX\bP_1 = \begin{bmatrix}
		\lambda_i &\bzero \\
		\bzero & \bY_1^\top \bX\bY_1
	\end{bmatrix} = \begin{bmatrix}
		\lambda_i &\bzero \\
		\bzero & \bB
	\end{bmatrix}. \qquad (\text{Let $\bB=\bY_1^\top \bX\bY_1$})
	$$
	As a result, $\bX$ and $\bP_1^\top\bX\bP_1$ are similar matrices such that they have same eigenvalues since $\bP_1$ is nonsingular (even orthogonal here, see Lemma~\ref{lemma:eigenvalue-similar-matrices}). We obtain 
	$$
	\det(\bP_1^\top\bX\bP_1 - \lambda\bI_n) =
	\footnote{By the fact that if matrix $\bM$ has a block formulation: $\bM=\begin{bmatrix}
			\bX & \bB \\
			\bC & \bD 
		\end{bmatrix}$, then $\det(\bM) = \det(\bX)\det(\bD-\bC\bX^{-1}\bB)$.
	}
	(\lambda_i - \lambda )\det(\bY_1^\top \bX\bY_1 - \lambda\bI_{n-1}).
	$$
	If $\lambda_i$ has multiplicity $k\geq 2$, then the term $(\lambda_i-\lambda)$ occurs $k$ times in the polynomial from the determinant $\det(\bP_1^\top\bX\bP_1 - \lambda\bI_n)$, i.e., the term occurs $k-1$ times in the polynomial from $\det(\bY_1^\top \bX\bY_1 - \lambda\bI_{n-1})$. In another word, $\det(\bY_1^\top \bX\bY_1 - \lambda_i\bI_{n-1})=0$ and $\lambda_i$ is an eigenvalue of $\bY_1^\top \bX\bY_1$. 

	Let $\bB=\bY_1^\top \bX\bY_1$. Since $\det(\bB-\lambda_i\bI_{n-1})=0$, the null space of $\bB-\lambda_i\bI_{n-1}$ is not none. Suppose $(\bB-\lambda_i\bI_{n-1})\bn = \bzero$, i.e., $\bB\bn=\lambda_i\bn$ and $\bn$ is an eigenvector of $\bB$. 
	
	From $
	\bP_1^\top\bX\bP_1 = \begin{bmatrix}
		\lambda_i &\bzero \\
		\bzero & \bB
	\end{bmatrix},
	$
	we have $
	\bX\bP_1 
	\begin{bmatrix}
		z \\
		\bn 
	\end{bmatrix} 
	= 
	\bP_1
	\begin{bmatrix}
		\lambda_i &\bzero \\
		\bzero & \bB
	\end{bmatrix}
	\begin{bmatrix}
		z \\
		\bn 
	\end{bmatrix}$, where $z$ is any scalar. From the left side of this equation, we have 
	\begin{equation}\label{equation:spectral-pro4-right}
		\begin{aligned}
			\bX\bP_1 
			\begin{bmatrix}
				z \\
				\bn 
			\end{bmatrix} 
			&=
			\begin{bmatrix}
				\lambda_i\bbeta_{i1}, \bX\bY_1
			\end{bmatrix}
			\begin{bmatrix}
				z \\
				\bn 
			\end{bmatrix} \\
			&=\lambda_iz\bbeta_{i1} + \bX\bY_1\bn.
		\end{aligned}
	\end{equation}
	And from the right side of the equation, we have 
	\begin{equation}\label{equation:spectral-pro4-left}
		\begin{aligned}
			\bP_1
			\begin{bmatrix}
				\lambda_i &\bzero \\
				\bzero & \bB
			\end{bmatrix}
			\begin{bmatrix}
				z \\
				\bn 
			\end{bmatrix}
			&=
			\begin{bmatrix}
				\bbeta_{i1} & \bY_1
			\end{bmatrix}
			\begin{bmatrix}
				\lambda_i &\bzero \\
				\bzero & \bB
			\end{bmatrix}
			\begin{bmatrix}
				z \\
				\bn 
			\end{bmatrix}\\
			&=
			\begin{bmatrix}
				\lambda_i\bbeta_{i1} & \bY_1\bB 
			\end{bmatrix}
			\begin{bmatrix}
				z \\
				\bn 
			\end{bmatrix}\\
			&= \lambda_i z \bbeta_{i1} + \bY_1\bB \bn \\
			&=\lambda_i z \bbeta_{i1} + \lambda_i \bY_1 \bn.  \qquad (\text{Since $\bB \bn=\lambda_i\bn$})\\
		\end{aligned}
	\end{equation}
	Combine Equation~\ref{equation:spectral-pro4-left} and Equation~\ref{equation:spectral-pro4-right}, we obtain 
	$$
	\bX\bY_1\bn = \lambda_i\bY_1 \bn,
	$$
	which means $\bY_1\bn$ is an eigenvector of $\bX$ corresponding to the eigenvalue $\lambda_i$ (same eigenvalue corresponding to $\bbeta_{i1}$). Since $\bY_1\bn$ is a combination of $\by_2, \by_3, \cdots, \by_n$ which are orthonormal to $\bbeta_{i1}$, the $\bY_1\bn$ can be chosen to be orthonormal to $\bbeta_{i1}$.
	
	To conclude, if we have one eigenvector $\bbeta_{i1}$ corresponding to $\lambda_i$ whose multiplicity is $k\geq 2$, we could construct the second eigenvector by choosing one vector from null space of $(\bB-\lambda_i\bI_{n-1})$ constructed above. Suppose now, we have constructed the second eigenvector $\bbeta_{i2}$ which is orthonormal to $\bbeta_{i1}$.  
	For such eigenvectors $\bbeta_{i1}, \bbeta_{i2}$, we can always find additional $n-2$ orthonormal vectors $\by_3, \by_4, \cdots, \by_n$ so that $\{\bbeta_{i1},\bbeta_{i2}, \by_3, \by_4, \cdots, \by_n\}$ forms an orthonormal basis in $\real^n$. Put the $\by_3, \by_4, \cdots, \by_n$ into matrix $\bY_2$ and $\{\bbeta_{i1},\bbeta_{i2},  \by_3, \by_4, \cdots, \by_n\}$ into matrix $\bP_2$:
	$$
	\bY_2=[\by_3, \by_4, \cdots, \by_n] \qquad \text{and} \qquad \bP_2=[\bbeta_{i1}, \bbeta_{i2},\bY_1].
	$$
	We then have
	$$
	\bP_2^\top\bX\bP_2 = 
	\begin{bmatrix}
		\lambda_i & 0 &\bzero \\
		0& \lambda_i &\bzero \\
		\bzero & \bzero & \bY_2^\top \bX\bY_2
	\end{bmatrix}
	=
	\begin{bmatrix}
		\lambda_i & 0 &\bzero \\
		0& \lambda_i &\bzero \\
		\bzero & \bzero & \bC
	\end{bmatrix},
	$$
	where $\bC=\bY_2^\top \bX\bY_2$ such that $\det(\bP_2^\top\bX\bP_2 - \lambda\bI_n) = (\lambda_i-\lambda)^2 \det(\bC - \lambda\bI_{n-2})$. If the multiplicity of $\lambda_i$ is $k\geq 3$, $\det(\bC - \lambda_i\bI_{n-2})=0$ and the null space of $\bC - \lambda_i\bI_{n-2}$ is not none so that we can still find a vector from null space of $\bC - \lambda_i\bI_{n-2}$ and $\bC\bn = \lambda_i \bn$. Now we can construct a vector $\begin{bmatrix}
		z_1 \\
		z_2\\
		\bn
	\end{bmatrix}\in \real^n $, where $z_1, z_2$ are any scalar values, such that 
	$$
	\bX\bP_2\begin{bmatrix}
		z_1 \\
		z_2\\
		\bn
	\end{bmatrix} = \bP_2 
	\begin{bmatrix}
		\lambda_i & 0 &\bzero \\
		0& \lambda_i &\bzero \\
		\bzero & \bzero & \bC
	\end{bmatrix}
	\begin{bmatrix}
		z_1 \\
		z_2\\
		\bn
	\end{bmatrix}.
	$$
	Similarly, from the left side of above equation we will get $\lambda_iz_1\bbeta_{i1} +\lambda_iz_2\bbeta_{i2}+\bX\bY_2\bn$. From the right side of above equation we will get $\lambda_iz_1\bbeta_{i1} +\lambda_i z_2\bbeta_{i2}+\lambda_i\bY_2\bn$. As a result, 
	$$
	\bX\bY_2\bn = \lambda_i\bY_2\bn,
	$$
	where $\bY_2\bn$ is an eigenvector of $\bX$ and orthogonal to $\bbeta_{i1}, \bbeta_{i2}$. And it is easy to construct the eigenvector to be orthonormal to the first two.

	The process can go on, and finally, we will find $k$ orthonormal eigenvectors corresponding to $\lambda_i$. 
	
	Actually, the dimension of the null space of $\bP_1^\top\bX\bP_1 -\lambda_i\bI_n$ is equal to the multiplicity $k$. It also follows that if the multiplicity of $\lambda_i$ is $k$, there cannot be more than $k$ orthogonal eigenvectors corresponding to $\lambda_i$. Otherwise, it will come to the conclusion that we could find more than $n$ orthogonal eigenvectors which leads to a contradiction.
\end{proof}

The proof of Theorem~\ref{theorem:spectral_theorem} is trivial from the lemmas above. Also, we can use Schur decomposition to prove the existence of spectral decomposition (see \citep{lu2021numerical}).

\section{Singular Value Decomposition (SVD)}\label{appendix:SVD}


\begin{svgraybox}
	\begin{theorem}[Reduced SVD for Rectangular Matrices]\label{theorem:reduced_svd_rectangular}
		For every real $n\times p$ matrix $\bX$ with rank $r$, then matrix $\bX$ can be factored as
		$$
		\bX = \bU \bSigma \bV^\top,
		$$ 
		where $\bSigma\in \real^{r\times r}$ is a diagonal matrix $\bSigma=diag(\sigma_1, \sigma_2 \cdots, \sigma_r)$ with $\sigma_1 \geq \sigma_2 \geq \cdots \geq \sigma_r$ and 
		
		$\bullet$ $\sigma_i$'s are the nonzero \textbf{singular values} of $\bX$, in the meantime, they are the (positive) square roots of the nonzero \textbf{eigenvalues} of $\trans{\bX} \bX$ and $ \bX \trans{\bX}$.
		
		$\bullet$ Columns of $\bU\in \real^{n\times r}$ contain the $r$ eigenvectors of $\bX\bX^\top$ corresponding to the $r$ nonzero eigenvalues of $\bX\bX^\top$. 
		
		$\bullet$ Columns of $\bV\in \real^{p\times r}$ contain the $r$ eigenvectors of $\bX^\top\bX$ corresponding to the $r$ nonzero eigenvalues of $\bX^\top\bX$. 
		
		$\bullet$ Moreover, the columns of $\bU$ and $\bV$ are called the \textbf{left and right singular vectors} of $\bX$, respectively. 
		
		$\bullet$ Further, the columns of $\bU$ and $\bV$ are orthonormal (by Spectral Theorem~\ref{theorem:spectral_theorem}). 
		
		In particular, we can write out the matrix decomposition $\bX = \bU \bSigma \bV^\top = \sum_{i=1}^r \sigma_i \bu_i \bv_i^\top$, which is a sum of $r$ rank-one matrices.
	\end{theorem}
\end{svgraybox}
If we append additional $n-r$ silent columns that are orthonormal to the $r$ eigenvectors of $\bX\bX^\top$, we will have an orthogonal matrix $\bU\in \real^{n\times n}$. Similar for the columns of $\bV$. 
We then illustrate the full SVD for matrices in Theorem~\ref{theorem:full_svd_rectangular} where we formulate the difference between reduced and full SVD in the blue text.

The comparison of reduced and full SVD is shown in Figure~\ref{fig:svd-comparison} where white entries are zero and blue entries are not necessarily zero.

To prove the existence of the SVD, we need to use the following lemmas. We mentioned that the singular values are the square roots of the eigenvalues of $\bX^\top\bX$. While, negative values do not have square roots such that the eigenvalues must be nonnegative.
\begin{svgraybox}
	\begin{lemma}[Nonnegative Eigenvalues of $\bX^\top \bX$]\label{lemma:nonneg-eigen-ata}
		For any matrix $\bX\in \real^{n\times p}$, $\bX^\top \bX$ has nonnegative eigenvalues.
	\end{lemma}
\end{svgraybox}
\begin{proof}[of Lemma~\ref{lemma:nonneg-eigen-ata}]
	For eigenvalue and its corresponding eigenvector $\lambda, \bbeta$ of $\bX^\top \bX$, we have
	$$
	\bX^\top \bX \bbeta = \lambda \bbeta \leadto \bbeta^\top \bX^\top \bX \bbeta = \lambda \bbeta^\top\bbeta. 
	$$
	Since $\bbeta^\top \bX^\top \bX \bbeta  = ||\bX \bbeta||^2 \geq 0$ and $\bbeta^\top\bbeta \geq 0$. We then have $\lambda \geq 0$.
\end{proof}

Since $\bX^\top\bX$ has nonnegative eigenvalues, we then can define the singular value $\sigma\geq 0$ of $\bX$ such that $\sigma^2$ is the eigenvalue of $\bX^\top\bX$, i.e., \fbox{$\bX^\top\bX \bv = \sigma^2 \bv$}. This is essential to SVD.

We have shown in Lemma~\ref{lemma:rankAB} that $rank$($\bX\bB$)$\leq$min($rank$($\bX$), $rank$($\bB$)).
However, the symmetric matrix $\bX^\top \bX$ is rather special that the rank of $\bX^\top \bX$ is equal to $rank(\bX)$. And the proof is provided in Lemma~\ref{lemma:rank-of-ata-x}.
%


In the form of SVD, we claimed the matrix $\bX$ is a sum of $r$ rank-one matrices where $r$ is the number of nonzero singular values. And the number of nonzero singular values is actually the rank of the matrix.

\begin{svgraybox}
	\begin{lemma}[The Number of Nonzero Singular Values Equals to the Rank]\label{lemma:rank-equal-singular}
		The number of nonzero singular values of matrix $\bX$ equals the rank of $\bX$.
	\end{lemma}
\end{svgraybox}
\begin{proof}[of Lemma~\ref{lemma:rank-equal-singular}]
	The rank of any symmetric matrix (here $\bX^\top\bX$) equals the number of nonzero eigenvalues (with repetitions) by Lemma~\ref{lemma:rank-of-symmetric}. So the number of nonzero singular values equals the rank of $\bX^\top \bX$. By Lemma~\ref{lemma:rank-of-ata-x}, $\bX^\top \bX$ and $\bX$ have same rank, so the number of nonzero singular values equals to the rank of $\bX$.
\end{proof}

We are now ready to prove the existence of SVD.
\begin{proof}[\textbf{of Theorem~\ref{theorem:reduced_svd_rectangular}: Existence of the SVD}]
	Since $\bX^\top \bX$ is a symmetric matrix, by Spectral Theorem~\ref{theorem:spectral_theorem} and Lemma~\ref{lemma:nonneg-eigen-ata}, there exists an orthogonal matrix $\bV$ such that
	
	$$
	\boxed{\bX^\top \bX = \bV \bSigma^2 \bV^\top},
	$$
	where $\bSigma$ is a diagonal matrix containing the singular values of $\bA$, i.e., $\bSigma^2$ contains the eigenvalues of $\bA^\top \bA$.
	Specifically, $\bSigma=diag(\sigma_1, \sigma_2 \cdots, \sigma_r)$ and $\{\sigma_1^2, \sigma_2^2, \cdots, \sigma_r^2\}$ are the nonzero eigenvalues of $\bA^\top \bA$ with $r$ being the rank of $\bA$. I.e., $\{\sigma_1, \cdots, \sigma_r\}$ are the singular values of $\bA$. In this case, $\bV\in \real^{p\times r}$.
	Now we are into the central part.
	\begin{mdframed}[hidealllines=\mdframehidelineNote,backgroundcolor=\mdframecolorSkip]
		Start from \fbox{$\bX^\top\bX \bv_i = \sigma_i^2 \bv_i$}, $\forall i \in \{1, 2, \cdots, r\}$, i.e., the eigenvector $\bv_i$ of $\bX^\top\bX$ corresponding to $\sigma_i^2$:
		
		1. Multiply both sides by $\bv_i^\top$:
		$$
		\bv_i^\top\bX^\top\bX \bv_i = \sigma_i^2 \bv_i^\top \bv_i \leadto ||\bX\bv_i||^2 = \sigma_i^2 \leadto ||\bX\bv_i||=\sigma_i.
		$$
		
		2. Multiply both sides by $\bX$:
		$$
		\bX\bX^\top\bX \bv_i = \sigma_i^2 \bX \bv_i \leadtosmall \bX\bX^\top \frac{\bX \bv_i }{\sigma_i}= \sigma_i^2 \frac{\bX \bv_i }{\sigma_i} \leadtosmall \bX\bX^\top \bu_i = \sigma_i^2 \bu_i,
		$$
		where we notice this form can find the eigenvector of $\bX\bX^\top$ corresponding to $\sigma_i^2$ which is $\bX \bv_i$. Since the length of $\bX \bv_i$ is $\sigma_i$, we then define $\bu_i = \frac{\bX \bv_i }{\sigma_i}$ with norm 1.
	\end{mdframed}
	These $\bu_i$'s are orthogonal because $(\bX\bv_i)^\top(\bX\bv_j)=\bv_i^\top\bX^\top\bX\bv_j=\sigma_j^2 \bv_i^\top\bv_j=0$. That is 
	$$
	\boxed{\bX \bX^\top = \bU \bSigma^2 \bU^\top}.
	$$
	Since $\bX\bv_i = \sigma_i\bu_i$, we have 
	$$
	[\bX\bv_1, \bX\bv_2, \cdots, \bX\bv_r] = [ \sigma_1\bu_1,  \sigma_2\bu_2, \cdots,  \sigma_r\bu_r]\leadto
	\bX\bV = \bU\bSigma,
	$$
	which completes the proof.
\end{proof}
By appending silent columns in $\bU$ and $\bV$, we can easily find the full SVD.

\newpage
\vskip 0.2in
\bibliography{bib}

\begin{thebibliography}{9}
\providecommand{\natexlab}[1]{#1}
\providecommand{\url}[1]{\texttt{#1}}
\expandafter\ifx\csname urlstyle\endcsname\relax
  \providecommand{\doi}[1]{doi: #1}\else
  \providecommand{\doi}{doi: \begingroup \urlstyle{rm}\Url}\fi

\bibitem[Bishop(2006)]{bishop2006pattern}
Christopher~M Bishop.
\newblock Pattern recognition.
\newblock \emph{Machine learning}, 128\penalty0 (9), 2006.

\bibitem[Lu(2021{\natexlab{a}})]{lu2021numerical}
Jun Lu.
\newblock Numerical matrix decomposition and its modern applications: A
  rigorous first course.
\newblock \emph{arXiv preprint arXiv:2107.02579}, 2021{\natexlab{a}}.

\bibitem[Lu(2021{\natexlab{b}})]{lu2021rigorous}
Jun Lu.
\newblock A rigorous introduction for linear models.
\newblock \emph{arXiv preprint arXiv:2105.04240}, 2021{\natexlab{b}}.

\bibitem[Rose(1982)]{rose1982linear}
Nicholas~J Rose.
\newblock Linear algebra and its applications (gilbert strang).
\newblock \emph{SIAM Review}, 24\penalty0 (4):\penalty0 499--501, 1982.

\bibitem[Strang(1993)]{strang1993fundamental}
Gilbert Strang.
\newblock The fundamental theorem of linear algebra.
\newblock \emph{The American Mathematical Monthly}, 100\penalty0 (9):\penalty0
  848--855, 1993.

\bibitem[Strang(2009)]{strang1993introduction}
Gilbert Strang.
\newblock \emph{Introduction to linear algebra}.
\newblock Wellesley-Cambridge Press Wellesley, 4th edition, 2009.

\bibitem[Strang(2019)]{strang2019linear}
Gilbert Strang.
\newblock \emph{Linear algebra and learning from data}.
\newblock Wellesley-Cambridge Press Cambridge, 2019.

\bibitem[Strang(2021)]{strang2021every}
Gilbert Strang.
\newblock \emph{Linear algebra for everyone}.
\newblock Wellesley-Cambridge Press Wellesley, 2021.

\bibitem[Trefethen and Bau~III(1997)]{trefethen1997numerical}
Lloyd~N Trefethen and David Bau~III.
\newblock \emph{Numerical linear algebra}, volume~50.
\newblock Siam, 1997.

\end{thebibliography}
\end{document}